\documentclass{article}

\usepackage[final, nonatbib]{neurips_2022}

\usepackage[utf8]{inputenc} % allow utf-8 input
\usepackage[T1]{fontenc}    % use 8-bit T1 fonts
\usepackage{hyperref}       % hyperlinks
\usepackage{url}            % simple URL typesetting
\usepackage{booktabs}       % professional-quality tables
\usepackage{amsfonts}       % blackboard math symbols
\usepackage{nicefrac}       % compact symbols for 1/2, etc.
\usepackage{microtype}      % microtypography
\usepackage{xcolor}         % colors

\usepackage{algorithm, algorithmic}
\usepackage[numbers]{natbib}
\usepackage{graphicx}
\usepackage{multicol}
\usepackage{amsmath}
\usepackage{amssymb}
\usepackage{mathtools}
\usepackage{amsthm}
\usepackage{wrapfig}

\theoremstyle{plain}

\theoremstyle{definition}

\theoremstyle{remark}

\usepackage{xspace}
\newcommand{\Algo}{\textsc{Dash}\xspace}

\DeclareMathOperator*{\argmax}{arg\,max}

\DeclareMathOperator{\diag}{\operatorname{diag}}
\DeclareMathOperator{\Op}{\operatorname{\bf Op}}

\newcommand{\BigO}{\mathcal O}
\DeclareMathOperator{\Conv}{\operatorname{\bf Conv}}

\DeclareMathOperator{\Id}{\operatorname{\bf Id}}
\DeclareMathOperator{\Zero}{\operatorname{\bf Zero}}
\DeclareMathOperator{\MaxP}{\operatorname{\bf MaxPool}}
\DeclareMathOperator{\AvgP}{\operatorname{\bf AvgPool}}
\DeclareMathOperator{\DilC}{\operatorname{\bf DilatedConv}}
\DeclareMathOperator{\AggC}{\operatorname{\bf AggConv}}

\usepackage{enumitem}
\usepackage{bm}
\def\*#1{\mathbf{#1}}

\usepackage{threeparttable}

\usepackage{multirow}
\usepackage[textsize=tiny]{todonotes}
\usepackage{subcaption}

\newcommand{\res}[2]{#1$\pm$#2}

\title{Efficient Architecture Search for Diverse Tasks}

\author{%
  Junhong Shen$^\ast$ \\
  Carnegie Mellon University\\
{}\texttt{junhongs@andrew.cmu.edu} \\
   \And
 Mikhail Khodak$^\ast$ \\
  Carnegie Mellon University\\
  \texttt{khodak@cmu.edu} \\
   \And
Ameet Talwalkar \\
  Carnegie Mellon University\\
  \texttt{talwalkar@cmu.edu} \\
}

\makeatletter
\def\thxfootnote{\xdef\@thefnmark{}\@footnotetext}
\makeatother

\begin{document}
\thxfootnote{$\ast$ Equal contribution.}

\maketitle

\begin{abstract}
    While neural architecture search (NAS) has enabled automated machine learning (AutoML) for well-researched areas, its application to tasks beyond computer vision is still under-explored. As less-studied domains are precisely those where we expect AutoML to have the greatest impact, in this work we study NAS for efficiently solving \textit{diverse} problems. 
	Seeking an approach that is fast, simple, and broadly applicable, we fix a standard convolutional network (CNN) topology and propose to search for the right kernel sizes and dilations its operations should take on. 
	This dramatically expands the model's capacity to extract features at multiple resolutions for different types of data while only requiring search over the operation space.
	To overcome the efficiency challenges of naive weight-sharing in this search space, we introduce \Algo, a differentiable NAS algorithm that computes the mixture-of-operations using the Fourier diagonalization of convolution, achieving both a better asymptotic complexity and an up-to-10x search time speedup in practice.
	We evaluate \Algo on ten tasks spanning a variety of application domains such as PDE solving, protein folding, and heart disease detection. \Algo outperforms state-of-the-art AutoML methods in aggregate, attaining the best-known automated performance on seven tasks. Meanwhile, on six of the ten tasks, the combined search and retraining time is less than 2x slower than simply training a CNN backbone that is far less accurate.
\end{abstract}
\vspace{5pt}

% !TEX root = main.tex

\section{Introduction}
\label{sec:intro}

The success of deep learning for computer vision and natural language processing has spurred growing interest in enabling similar breakthroughs for other domains such as biology, healthcare, and physical sciences. Consequently, there is enormous potential for neural architecture search to help automate model development in these diverse areas. However, while extensive NAS research has been devoted to improving the search speed \citep{pham2018enas,liu2019darts} and automatically attaining state-of-the-art performance on vision datasets such as CIFAR and ImageNet \citep{li2021gaea}, the resulting algorithms have subpar performance beyond the tasks on which they were developed.
For example, in the analysis of NAS-Bench-360 \citep{tu2021nb360}, a recent benchmark designed for improving task diversity in NAS evaluation, the authors showed a significant gap between models found by NAS methods, such as DARTS \citep{liu2019darts} and DenseNAS \citep{fang2020densenas}, and hand-crafted expert architectures on a number of distinct applications.

\begin{wrapfigure}{r}{0.524\textwidth}
  \centering
   \begin{subfigure}[b]{0.31\textwidth}
        \includegraphics[width=\textwidth]{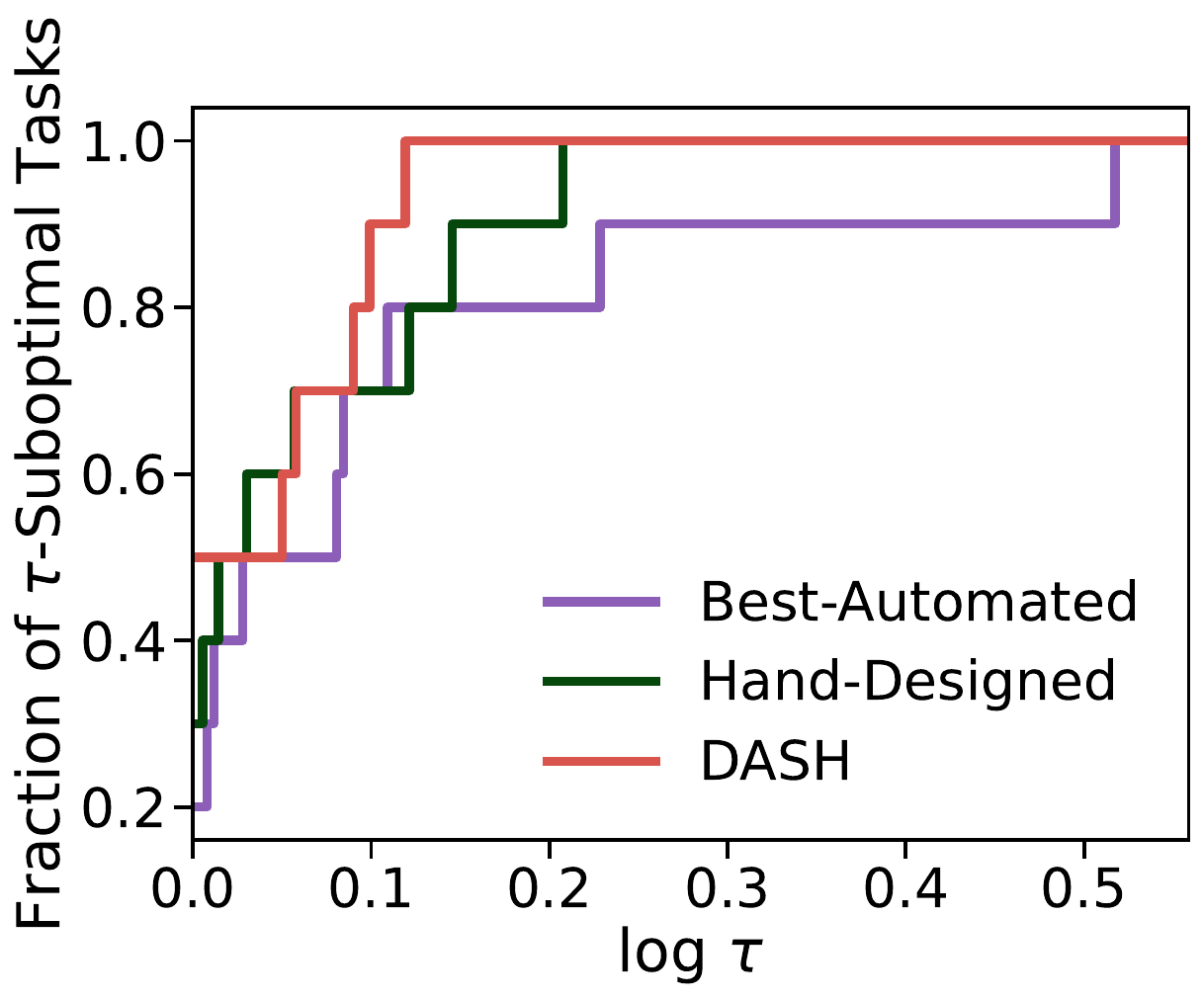}
        \subcaption{}
        \label{fig:introa}
    \end{subfigure}~
    \begin{subfigure}[b]{0.2\textwidth}    
        \includegraphics[width=\textwidth]{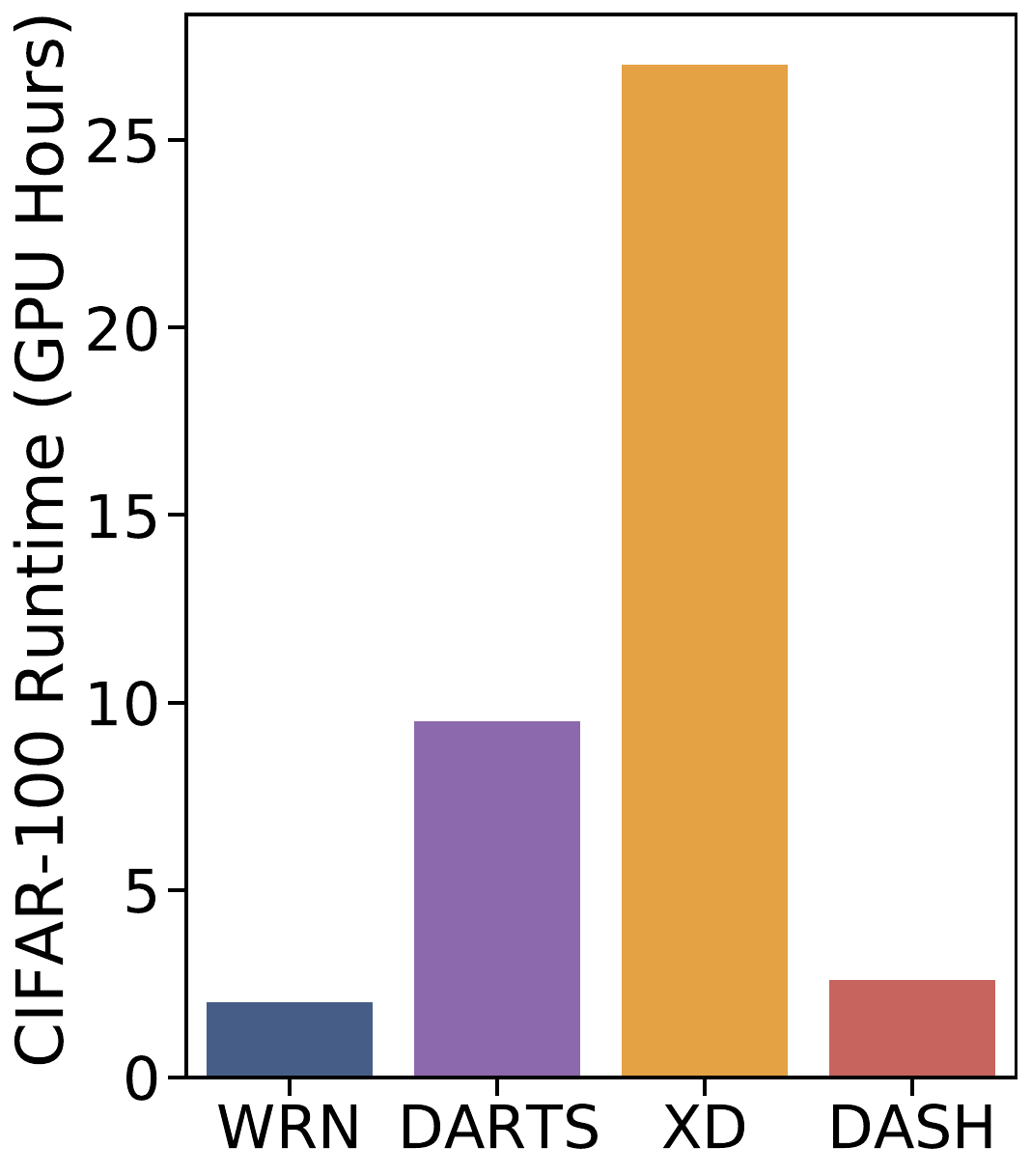}
        \subcaption{}
        \label{fig:introb}    
    \end{subfigure}
 \vspace{-5mm}
  \caption{\small (a) Comparing the aggregate performance of the best AutoML methods (task-wise), hand-designed models, and \Algo on ten diverse tasks via performance profiles (defined in Section~\ref{sec:exp_setup}). Larger values (larger fractions of tasks on which a method is within $\log\tau$-factor of the best) are better. (b) Runtime for Wide ResNet, DARTS, XD, and \Algo on CIFAR-100. XD is too expensive to be applied to other tasks considered in this work \citep{tu2021nb360, roberts2021xd}.\looseness-1}
	 \label{fig:intro}
	\vspace{-1mm}
\end{wrapfigure}

To improve the generalizability of AutoML methods,  recent works such as AutoML-Zero \citep{real2020automlzero} and XD-operations \citep{roberts2021xd} propose to relax the inductive biases encoded in the standard search spaces. 
However, the former is not designed for practical deployment, and the latter is too expensive to execute even for simple problems like CIFAR-100 (Fig.~\ref{fig:introb}). 
Hence we ask: 
\emph{is there an approach that can provide sufficient expressivity  to yield high accuracy across multiple domains, while still retaining the faster search and the more efficient final models of discrete architecture search?}

In this work, we answer in the affirmative by introducing a novel NAS method called \Algo (\textbf{D}iverse-task \textbf{A}rchitecture \textbf{S}earc\textbf{H}). In order to attain \textit{multi-domain capability}, \Algo adapts standard CNN backbones to various learning problems by finding substitutes for their layer operations. 
Concretely, we consider a search space of cross-scale dilated convolutions which are effective for multi-scale feature extraction~\citep{WaveNet, Yang2017ImprovedVA} and context aggregation~\citep{Yu2016MultiScaleCA, Chen2018DeepLabSI}. Our key difference from past search spaces is that we explicitly consider filters with \textit{a wide range of kernel sizes and dilations}---while most NAS methods only handle kernels with maximum size $5$ and dilation rate $2$, our proposed operator space includes not only the conventional small kernels but also significantly larger ones with size $15$ or dilation $127$ (Fig.~\ref{fig:time1}). This design choice is motivated by the fact that large kernels can capture input relations for dense prediction problems~\citep{Peng2017LargeKM}, model long-range dependencies for sequence tasks~\citep{bai2018tcn,bai2019trellis}, and resemble global-attention in Transformers~\citep{Liu2022ACF}. Thus, \Algo's cross-scale  search space enables adaptation  to diverse downstream tasks, unlike prior NAS work which targets image classification and assumes that small kernels are sufficient.

However, \textit{efficiently} searching for an appropriate kernel configuration in this expansive cross-scale search space is non-trivial. Indeed, for existing NAS algorithms, the cost of exploring a combinatorially large set of operators is substantial. Even for weight-sharing methods that are known for efficiency, e.g., DARTS \citep{liu2018darts},  the computational complexity scales directly with the number of kernels considered and quadratically with the largest kernel size. To overcome this obstacle, \Algo explores multi-scale convolutions via three techniques---the first two exploit mathematical properties of convolutions, and the last one takes advantage of fast matrix multiplication on GPUs. Specifically:
\begin{enumerate}[leftmargin=*,noitemsep]\setlength\itemsep{5pt}
	\item Using the {\bf linearity} of convolutions, we mix several convolutions by computing one convolution equipped with a combined kernel rather than applying each filter separately and aggregating multiple outputs. While the number of convolution computations required by the naive aggregation of $|K|$ possible kernel sizes and $|D|$ possible dilations is $\BigO(|K||D|)$, our approach has $\BigO(1)$ complexity, independent of the search space size.
	
	\item Using the {\bf diagonalization} of convolutions, we relegate a major portion of the computation to element-wise multiplication in the Fourier domain, minimizing the effect of the largest kernel size on the  complexity of our algorithm.
	For instance, a standard 1D convolution requires $\BigO(nk)$ operations to convolve a size-$k$ kernel  with a length-$n$ input, but a Fourier convolution takes only $\BigO(n\log n)$, a critical improvement that makes searching over large kernels significantly easier.
	
	\item Our final strategy is to use Kronecker products of undilated kernels and small sparse matrices to compute dilated kernels quickly on GPUs. This brings an additional two-fold speedup on top of the previous techniques.
\end{enumerate}
Aside from these innovations, \Algo employs the standard weight-sharing scheme of training a supernet, discretizing to obtain a model, and retraining the  model for end tasks \citep{liu2019darts}. We analyze the asymptotic complexity of the first two techniques and verify the practical utility when all three are combined together. In particular, \Algo achieves a ten-fold speedup in total for differentiable NAS over the multi-scale search space. Moreover, we show that  searching over large kernels is necessary to solve diverse problems and that each technique on its own cannot scale in this large-kernel setting.

In terms of accuracy performance, we evaluate \Algo on ten datasets spanning multiple application areas such as PDE solving, protein folding, and disease prediction from NAS-Bench-360~\citep{tu2021nb360}.
As shown in Fig.~\ref{fig:intro}, \Algo yields models with better aggregate performance than those returned by leading AutoML methods as well as hand-designed task-specific architectures. As for individual tasks, \Algo beats all past automated approaches on seven of the ten problems and exceeds hand-designed models on seven, simultaneously maintaining strong efficiency relative to  weight-sharing methods like DARTS. The empirical success of \Algo implies that CNNs with appropriate kernels can be  competitive baselines for problems where expert architectures are not available. Our code is made public at \url{https://github.com/sjunhongshen/DASH}.
% !TEX root = main.tex

\section{Related Work}
\label{sec:related_work}

Neural architecture search (NAS) aims to automate the design of neural networks. Recently, there has been significant progress in both search space design~\citep{zoph2018nas, liu2019darts,roberts2021xd} and search strategy development~\citep{pham2018enas,xie2019snas, liu2019darts,xu2020pcdarts,li2021gaea}. However, these  methods are mostly evaluated on image classification or segmentation, with a few focusing on new applications such as image restoration~\citep{suganumaICML2018}, audio classification~\citep{Mehrotra2021NASBenchASRRN}, and machine translation~\citep{So2019TheET}. Problems beyond the vision and language domains are less-explored. 

In this work, we seek to improve the generalizability of NAS by a morphism-based approach~\citep{Wei2016NetworkM, Chen2016Net2NetAL,roberts2021xd} that adapts existing CNN backbones to target tasks.  We specifically focus on expanding the operation space to multiple types of convolutions, simultaneously varying the kernel size and the dilation factor. Past work in related directions has at most studied the easier problem of altering dilation alone, and only for vision tasks \citep{chen2018searching}. Therefore, although convolution has been an integral part of NAS, how to search over a large set of convolutional operators remains an open problem. In the following, we identify four types of solutions from existing work and illustrate their limitations. 

\paragraph{Differentiable Architecture Search (DARTS)} We can treat convolutions as ordinary operators and apply a scalable NAS algorithm. In particular, DARTS \citep{liu2019darts} introduces continuous relaxation to the weight-sharing paradigm \citep{pham2018enas} and allows us to gain information about many networks efficiently by training a combined supernet. The algorithm relaxes the discrete set of operations at each edge in a computational graph as a softmax so the search process is end-to-end differentiable and amenable to regular optimizers. After search, it discretizes the weights to output a valid architecture. The original DARTS search space contains only  four convolutions with  kernel sizes  no larger than $5$ and dilation rates no larger than $2$. While this small search space might be enough for low-resolution image input, it is insufficient for diverse tasks such as high-dimensional time series problems~\citep{tu2021nb360}. Although one could add more convolutions one-by-one to the operator set to augment performance, this approach scales poorly, as reflected in the limited search spaces of similar methods like AMBER \citep{zhang2021amber}. In fact, AMBER has to shift the kernel size up to achieve good performance on long-sequence genomic data.

\paragraph{MergeNAS and RepVGG} An alternative way to explore the convolutional search space is to take advantage of the operator's linearity. That is,  we can first mix the kernels and then apply convolution once, unlike DARTS which computes each convolution separately and outputs the aggregated result to the next layer. This kernel-mixing strategy, which we call \textsl{mixed-weights} and will  formally define in Section~\ref{sec:methods2}, has been employed by MergeNAS \citep{Wang2020MergeNASMO} and RepVGG \citep{Ding2021RepVGGMV} to improve architecture search robustness and VGG inference speed, respectively. It works well for \textit{a few small} kernels. However, we will show later that similar to DARTS, \textsl{mixed-weights} on its own is also insufficient for searching over \textit{a diverse set of large kernels} which is crucial to solving a wide range of problems. 

\paragraph{Expressive Diagonalization (XD)} Apart from linearity, XD-operations \citep{roberts2021xd} propose to utilize the convolutional theorem in architecture search. In particular, XD expresses the convolution acting on input $\*x$ with filter $\*w$ as $\*K\diag(\*L\*w)\*M\*x$, where $\*K,\*L,\*M$ are appropriate discrete Fourier transforms, and constructs an expansive search space  by replacing these transforms with  searched kaleidoscope matrices \citep{dao2020kaleidoscope}. Although this new  search space includes all types of convolutions, the  search process is unacceptably long  even for simple benchmarking tasks such as CIFAR-100 (Fig.~\ref{fig:introb}), let alone the more complex set of diverse problems that we consider in this paper. In addition, the output architectures of XD are as inefficient as the supernet due to the absence of a discretization step. 

\paragraph{Single-Path NAS} Lastly, Single-Path NAS~\citep{Stamoulis2019SinglePathND} defines a large filter and uses its subsets for smaller filters. This DARTS-based method compensates operator heterogeneity for efficiency during search. It does not handle search spaces with many large kernels and is not evaluated on diverse tasks.

Outside the field of AutoML, there is also emerging interest in designing general-purpose models such as Perceiver IO \citep{jaegle2107} and Frozen Pretrained Transformer \citep{Lu2021PretrainedTA}. However, these Transformer-based models do not adapt the network to the target tasks and are generally harder to train compared with CNNs. In Table~\ref{table:acc}, we evaluate Perceiver IO and show that its performance is not 
ideal.
% !TEX root = main.tex

\section{Methods}
\label{sec:methods}

\begin{algorithm}[tb]
  \caption{\Algo}
  \label{alg:}
\begin{algorithmic}
  \STATE {\bfseries Input:} training data $Z$, loss function $l$, the set of kernel sizes $K$, the set of dilation rates $D$, and subsampling ratio $p$
  \STATE Initialize the backbone and replace each $\Conv$ layer with the mixed operation $\AggC_{K,D}$
  \WHILE{not converged}
  \STATE Subsample $p|Z|$ training points uniformly at random
  \STATE Compute forward pass using Equation \ref{equ:fastxd}
  \STATE Descend the architecture parameters $\alpha$ by $\nabla_\alpha l(\*w, \alpha)$ and the model weights $\*w$ by $\nabla_{\*w} l(\*w, \alpha)$
  \ENDWHILE
  \STATE Select $\argmax_{k\in K, d\in D}\alpha_{k, d}$ for each $\AggC$ layer
  \STATE Tune retraining hyperparameters on a validation subset of the training data
  \STATE Retrain the discretized model with all training data
\end{algorithmic}
\end{algorithm}

Now, we describe the details of \Algo (Algorithm \ref{alg:}). We first explain how \Algo leverages existing networks to initialize the supernet and generate different models for diverse tasks. Then, we formally define the multi-scale convolution search space and propose a fast way to search this space using the three efficiency-motivated techniques mentioned earlier. Finally, we outline the procedure for discretizing the search space and retraining the searched model.

\subsection{Decoupling Topology and Operations}
\label{sec:methods1}
Every architecture is a mapping from model weights to functions and can be described by a directed acyclic graph $G(V, E)$. Each edge in $E$ is characterized by $(u, v, \Op)$, where $u, v \in V$ are nodes and $\Op$ is an operation applied to $u$. Node $v$ aggregates the outputs of its incoming edges. NAS aims to automatically select the edge operations and the graph topology to optimize some objective. For each edge, $\Op$ is chosen from a search space $S = \{\Op_a | a \in A\}$ where $a\in A$ are architecture parameters. In past work, $A$ usually indexes a small set of operations. For instance, the DARTS search space specifies $A_{discrete} = \{1, \dots, 8\}$ with $S =\{\Zero$, $\Id$, $\MaxP_{3\times3}$, $\AvgP_{3\times3}$, $\Conv_{3\times3}$, $\Conv_{5\times5}$, $\DilC_{3\times3,2}$, $\DilC_{5\times5,2}\}$. However, $A$ can also be defined systematically to identify operator properties, e.g.,  $\{($kernel size $k$, dilation $d)\}$ for convolutions.

A common way to determine the network topology is to search for blocks of operations and stack several blocks together. In this work, we take a different, morphism-based approach \citep{Wei2016NetworkM, Chen2016Net2NetAL, roberts2021xd}: we use existing networks as backbones and replace certain layers in the backbone with the searched operations. Specifically, we select a set of architectures to accommodate both 2D and 1D datasets. Convolutional layers with different kernels can then be plugged into these networks. An advantage of decoupling topology and operation search is flexibility: the searched operators can vary from the beginning to the end of a network, so features at different granularities can be processed differently.

We pick Wide ResNets (WRNs) \citep{zagoruyko2016wideresnet, IsmailFawaz2020InceptionTimeFA} as the backbone networks due to their simplicity and effectiveness in image and sequence modeling. Before search, the supernet is initialized to the backbone. Then,  all $\Conv$ layers are substituted with an operator $\AggC_{K, D}$ (short for aggregated convolution) that represents the new search space which we now define. For simplicity, our mathematical discussion will stick to the 1D case, though our experiments are on both 1D and 2D data.

\subsection{Efficiently Searching for Multi-Scale Convolutions}
\label{sec:methods2}
A convolution filter is specified by the kernel size $k$ and the dilation rate $d$ (we do not consider stride which does not change the filter shape). The effective filter size is $(k-1)d + 1$ with nonzero entries separated by $d-1$ zeros. Let $\Conv_{k,d}$ be the convolution with kernel size $k$, dilation rate $d$, $c_{in}$ input channels, and $c_{out}$ output channels. Given input data with shape $n$, let $K$ be our interested set of kernel sizes, $D$ the set of dilations. We define the $\AggC_{K, D}$ search space as
\begin{align}
  S_{\AggC_{K, D}} = \{\Conv_{k,d} | k \in K, d\in D\}.
  \label{equ:Saggconv}
\end{align}
Hence, $A = K \times D$ in previous notations. $S_{\AggC_{K, D}}$ contains a collection of convolutions with receptive field size ranging from $k_{min}$ to $d_{\max} (k_{\max} -1)+1$, which allows us to discover models that process the input with varying resolution  at each layer. 

To retain the efficiency of discrete NAS, we apply the continuous relaxation scheme of DARTS to $S_{\AggC_{K, D}}$, which mixes all operations in the space using architecture parameters $\{\alpha_{k, d}\in\triangle_{|K||D|} | (k, d) \in K\times D\}$\footnote{$\triangle$ denotes the probability simplex.}, so the output of each edge in the computational graph is
\begin{align}
\begin{split}
&\AggC_{K,D} (\*x) := \sum_{k\in K}\sum_{d\in D} \alpha_{k,d}\cdot \Conv(\*w_{k,d})(\*x).
\label{equ:mixed-results}
\end{split}
\end{align}
Here $\{\*w_{k, d} | (k, d) \in K\times D\}$ are the kernel weights. The resulting supernet can be trained end-to-end, and our hope is that after search, the most important operation is assigned the highest weight. However, the complexity of computing the above summation directly, a baseline algorithmic approach we call \textbf{\textsl{mixed-results}}, is $\BigO(c_{in}c_{out}(|K||D|+\bar K)n)$, where $\bar K:=|D|\sum_{k\in K}k$. \textsl{Mixed-results} can be  expensive when we increase the maximum element in $K$ or $D$ with larger input size $n$. To improve upon it, we propose three techniques which build up to the efficiency-oriented \Algo.

\begin{table}[t]
	\centering
	\caption{\small Complexity of different methods for computing $\AggC$. For notation, $\Bar{K} := |D| \cdot \sum_{k\in K} k$, $\bar D := \max_{k, d} (k-1)d+1$. A detailed analysis is provided in Appendix~\ref{sec:appendix:analysis}.}
	\vspace{0.1cm}
  \resizebox{\textwidth}{!}{
  \begin{tabular}{ccc}
		\toprule
		Method & MULTs & ADDs \\
		\midrule
		\textsl{mixed-results} (Eqn.~\ref{equ:mixed-results}) &$(c_{in}c_{out}\Bar{K} + c_{out}|K||D|)n$&$(c_{in}c_{out}\Bar{K} + c_{out}|K||D|)n$\\
		\textsl{mixed-weights} (Eqn.~\ref{equ:mixed-weights}) &$c_{in}c_{out}(\Bar{K}+\bar Dn)$&$c_{in}c_{out}\bar D(|K||D|+n)$ \\
		\Algo (Eqn.~\ref{equ:fastxd}) & $c_{in}c_{out}(\Bar{K}+n)+\BigO(c_{in}c_{out}n\log n)$ &$c_{in}c_{out}(|K||D|\bar D+n)+\BigO(c_{in}c_{out}n\log n)$\\
		\bottomrule
	\end{tabular}
	}
	\label{table:complexity}
\end{table}

\subsubsection{Technique 1: Mixed-Weights}
Since convolution is linear, instead of computing $|K||D|$ convolutions, we can combine the kernels and compute convolution once. We call this approach \textbf{\textsl{mixed-weights}}: 
\begin{align}
\begin{split}
  &\AggC_{K,D}(\*x)=\Conv\left(\sum_{k\in K}\sum_{d\in D} \alpha_{k,d}\cdot \*w'_{k,d}\right)(\*x).
  \label{equ:mixed-weights}
\end{split}
\end{align}
Here $\*w'$ is the properly padded  version of $\*w$ (appending $0$'s at the end of each dimension) so that filters of different sizes can be added. The aggregated kernel has size $\bar D:=\max_{k, d} (k-1)d+1$ and the $n$-dependent term of the complexity of \textsl{mixed-weights} is $c_{in}c_{out}\bar Dn$. Hence, it removes the direct dependence of the leading-order term on $|K||D|$,  the search space size, that \textsl{mixed-results} had.

\subsubsection{Technique 2: Fourier Convolution}
If we wish to increase the kernel size and dilation with the input size, the complexity of \textsl{mixed-weights} will still grow {\em implicitly} with the search space size through the dependence on $\bar D$. To address this issue, we combine the kernel re-weighting idea with another technique motivated by the convolution theorem. Given a kernel $\*w$, recall that $\Conv(\*w)(\*x) =\*F^{-1}\diag(\*F\*w')\*F\*x$, where $\*F$ is the discrete Fourier transform (DFT) and $\diag(\*z)$ is a diagonal matrix with entries $\*z$. In other words, convolution involves multiplying the DFT of the kernel by that of the input. Since the DFT can be applied in time $\BigO(n\log n)$ using the Fast Fourier Transform (FFT) and apart from that we only need element-wise multiplication, this yields an efficient approach to reducing dependence on the combined kernel size $\bar D$. While \citet{roberts2021xd} replaced the DFTs with a \textit{continuous} set of matrices for XD-operations, our approach can be viewed as replacing the middle DFT with a  \textit{discrete} set of matrices for the purpose of efficiency. Accordingly, \Algo computes $\AggC_{K,D}$ as follows: 
\begin{align}
\begin{split}
  &\AggC_{K,D}(\*x) = \*F^{-1}\diag\left(\*F\left(\sum_{k\in K}\sum_{d\in D} \alpha_{k,d}\cdot \*w'_{k,d}\right)\right)\*F\*x.
  \label{equ:fastxd}
\end{split}
\end{align}
Note that while the kernel changes for each $\Conv_{k,d}$, the input does not. Hence, we also save time by transforming the input to the frequency domain only once.

\begin{multicols}{2}
\begin{figure}[H]
  \centering
 \includegraphics[width=0.44\textwidth]{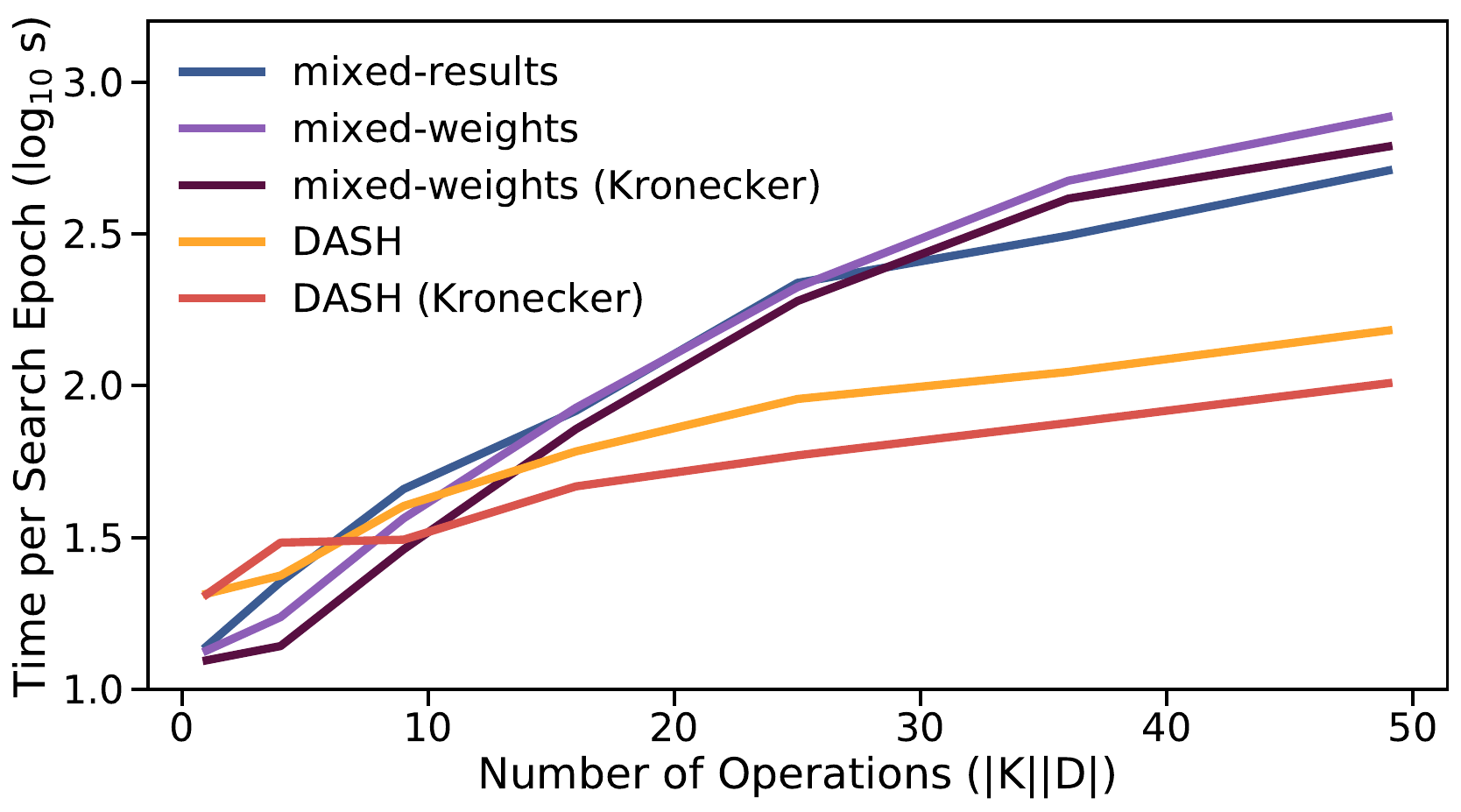}
  \caption{\small $\log_{10}$ time for one search epoch vs. number of ops in $S_{\AggC_{K, D}}$. We vary the search space by letting $K = \{2p+1 | 1\leq p \leq c\}$, $D = \{2^q-1 | 1\leq q\leq c\}$ and increasing $c$ from $1$ to $7$.}
\label{fig:time1}
\end{figure}

\begin{figure}[H]
  \centering
  \includegraphics[width=0.44\textwidth]{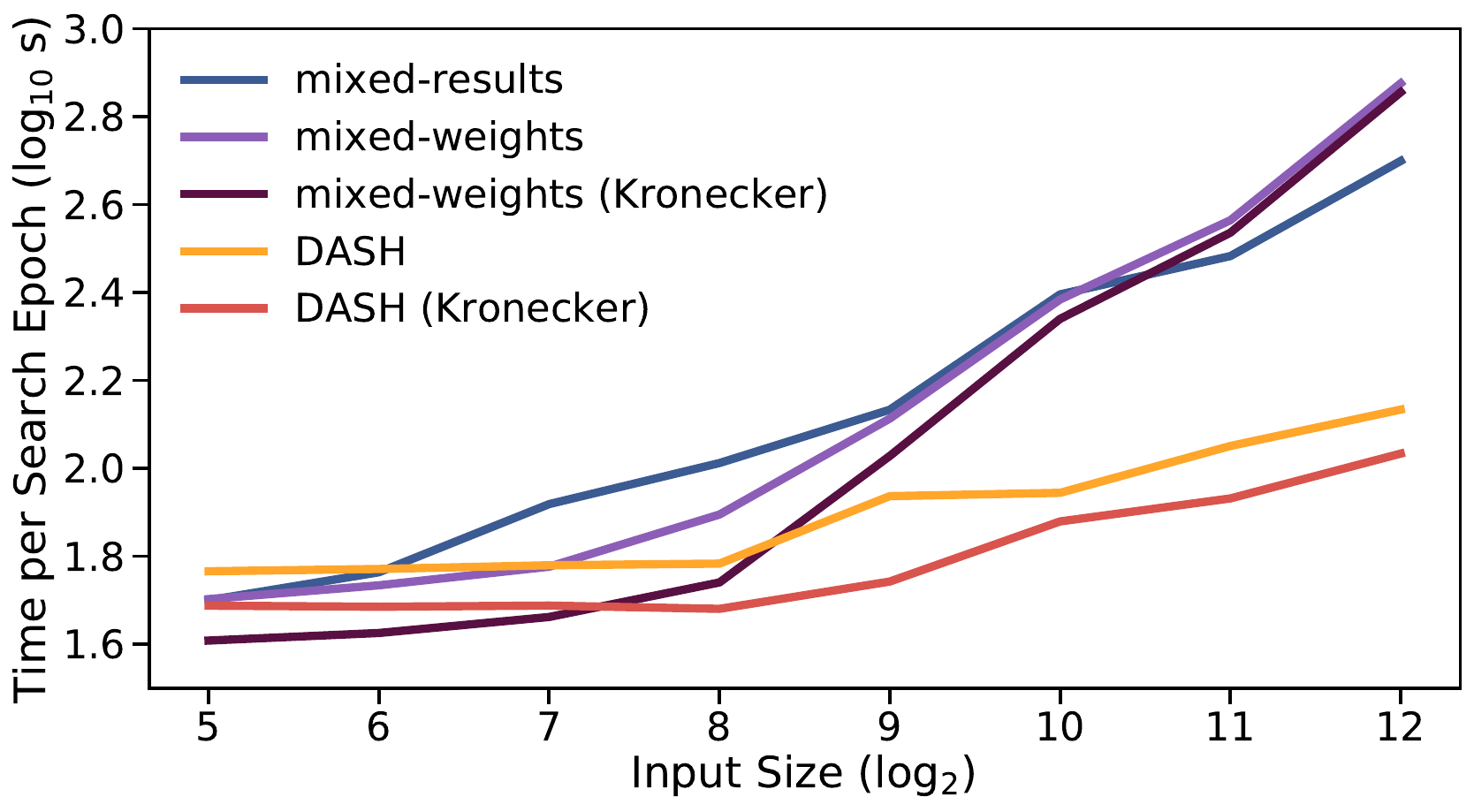}
  \caption{\small $\log_{10}$ time for one search epoch vs. input length of single-channel 1D data. We fix $K = \{3, 5, 7, 9, 11\}$, $D = \{1, 3, 7, 15, 31\}$ and test $n \in \{2^5, \dots, 2^{12}\}$.}
\label{fig:time2}
\end{figure}
\end{multicols}

In Table \ref{table:complexity}, we report the theoretical complexities of the baselines and \Algo, the latter leveraging both Technique 1 and 2. It is easy to obtain the operation complexities of \textsl{mixed-weights} and \textsl{mixed-results}. For \Algo, the number of multiplications and additions can be attributed to the inner weight sum and multi-channel product (the first term) as well as three FFTs (the second). A detailed analysis is provided in Appendix~\ref{sec:appendix:analysis}. We see that \textsl{mixed-weights} is favorable to \textsl{mixed-results} when $\bar D<\Bar{K}=\BigO(|D|k_{\max}^2)$, which occurs with large kernels and a few dilations. On the other hand, only \Algo completely separates the dominant terms containing $c_{in}c_{out}n$ from the size of the search space and its elements, replacing them by $\BigO(\log n)$, which is small for any realistic $n$. As we increase $k_{\max}$ and $d_{\max}$ for larger inputs, this will also lead to a slower asymptotic increase in complexity, making \Algo an attractive choice for the multi-scale search space where $\bar D$ is large by design to extract possible long-range dependencies in the data.

\subsubsection{Technique 3: Kronecker Dilation}
To efficiently implement the kernel summation in \textsl{mixed-weights} and \Algo on a GPU, we introduce our final technique: after initializing $\*w_{k,d}$ for each $\Conv_{k, d}$ separately, we use a Kronecker product $\otimes$ to transform the undilated kernels into dilated forms. For example, to compute a 2D convolution with dilation $d$, we introduce the sparse pattern matrix $\*P \in \mathbb{R}^{d \times d}$ whose entries are all $0$'s except for the upper-left entry $\*P_{1,1}=1$:
\begin{align}
\small
\*P = 
    \begin{bmatrix}
1 & 0 & \cdots & 0\\
0 & 0 & \cdots & 0\\
\vdots&\vdots&\ddots&\vdots\\
0 & 0 & \cdots & 0
\end{bmatrix}.
\end{align}

Then, $\*w_{k,d} = \*w_{k, 1} \otimes \*P$. Beyond the theoretical gains shown in \citet{Wu2019TreeStructuredKC}, this dilation strategy is empirically faster than the standard way of padding $0$'s into $\*w_{k, 1}$ (Fig.~\ref{fig:time1} and~\ref{fig:time2}). After dilating the kernels, we sum them together, zero-pad to match the input size, and apply the FFTs.

\subsubsection{Ablation Study for the Proposed Techniques}
To check that our asymptotic analysis leads to actual speedups and perform an ablation study on the proposed techniques, we evaluate the three methods on single-channel 1D input (experiment details are in Appendix~\ref{sec:appendix:mnist}). Since both \textsl{mixed-weights} and \Algo require kernel summation, which can be implemented with Kronecker dilation (Technique~3), we compare five methods in total.

Fig.~\ref{fig:time1} illustrates the combined forward- and backward-pass time in log scale for one search epoch vs. the size of $S_{\AggC_{K,D}}$ when $n = 1000$. For small $\bar D$, the FFT overhead makes \Algo runtime slightly longer but the difference is negligible. However, as $\bar D$ increases, the \Algo curves grow much slower whereas the runtimes for the other methods scale with the number of operations. In Fig.~\ref{fig:time2}, we fix $K$ and $D$ to study how runtime is affected by input size $n$. Both \textsl{mixed-results} and \textsl{mixed-weights} become extremely inefficient for large $n$'s which commonly occur in time-series or signal processing. Surprisingly, \Algo's runtime does not increase much with $n$. We hypothesize that this is due to wallclock-time being dominated by data-passing at that speed.

In general, Technique~1 on its own  scales poorly for the considered search space. This is why methods like MergeNAS \citep{Wang2020MergeNASMO} cannot be used in our setting. Though XD makes use of Technique~2, it considers a parametrized space with infinitely many operations that need to be continuously evolved and is too expensive to be applied to tasks beyond CIFAR-100 (c.f. Fig.~\ref{fig:introb}). Technique~3 contributes to 2$\times$ speedups for both \textsl{mixed-weights} and \Algo. Overall, \Algo (Kronecker)  leads to about 10$\times$ search-time speedups compared to the \textsl{mixed-results} scheme of DARTS for both the large operation space and large input size regimes. Hence, we use this version of \Algo in later experiments.

\subsection{The Full Pipeline: Architecture Search, Hyperparameter Optimization, and Retraining}
\label{sec:methodslast}

Having shown the main techniques for searching a large space of kernel patterns, we now detail  the full search and model development pipeline. Given a dataset, we set $K = \{3 + d(p-1) | 1\leq p \leq p_{\max}\}$ for kernel sizes and $D = \{2^q-1 | 1\leq q\leq q_{\max}\}$ for dilations. For 2D input, we set $d$ to 2, $p_{\max}$ to $4$, and $q_{\max}$ to 4. For longer 1D sequence data, we set $d = 4$, $p_{\max} = 5$, and  $q_{\max} = 4$. For instance, CIFAR has size $3 \times 32 \times 32$ where $3$ is the number of channels and $n = 32$. The corresponding $K$ is $\{3, 5, 7, 9\}$ and $D$ is $\{1, 3, 7, 15\}$. To normalize architecture parameters into a probability distribution, we adopt the soft Gumbel Softmax activation, similar to \citet{xie2019snas}. 

The backbone networks are as follows. For 2D tasks, we use WRN 16-1 as the search backbone to accelerate supernet training and WRN 16-4 for retraining. For 1D tasks, we use 1D WRN~\citep{IsmailFawaz2020InceptionTimeFA} in the entire pipeline. During search, we subsample the training data at each epoch. Given the loss for the target task, \Algo jointly optimizes the model weights and the architecture parameters using direct gradient descent. This single-level optimization is more efficient than two-stage NAS, which finds initial assignments for architecture parameters and trains the candidates from scratch.

After searching for a predefined number of epochs, we discretize the search space and pick $\Conv_{k, d} \in S_{\AggC_{K, D}}$ with the largest weight for each layer. The final model has a similar overall structure to the backbone, but the intermediate operations are tailored to the target task. To improve training stability, we additionally add a simple hyperparameter tuning stage between search and retraining using grid search (configuration space shown in  Appendix~\ref{appendix:hyperparamconfig}).

For each  setting, we train the discretized model on \textit{a subset} of the training data for \textit{fewer} epochs so the tuning cost is a small fraction of the entire pipeline's cost (Table~\ref{table:timebreakdown}), Then, we evaluate the performance on a holdout validation set and select the configuration with the best validation score. As a final step, we retrain the discretized model with the optimal hyperparameters on all training data until convergence. Like other weight-sharing methods with discretization, our final model will be more efficient than the supernet.

% !TEX root = main.tex

\section{Evaluation}
\label{sec:eval}

We evaluate the performance of \Algo on diverse tasks using ten datasets from NAS-Bench-360 \citep{tu2021nb360}, a benchmark spanning multiple application domains, input dimensions, and learning objectives.\footnote{For completeness, we give a task summary in the Appendix.} These include classical vision tasks such as CIFAR-100 where CNNs do well, scientific computing tasks such as Darcy Flow where standard CNN backbones can perform poorly \citep{roberts2021xd, li2021fno}, sequence tasks such as DeepSEA where large dilations are preferred \citep{bai2018tcn,zhang2021amber}, and many others.
Thus, our evaluation will not only test whether \Algo can find good architectures in the proposed new search space, but also investigate whether multi-scale convolution is a strong competitor for solving different problems. In fact, our results show that \Algo is a top choice for many tasks, obtaining in-aggregate the best speed-accuracy trade-offs among the methods we evaluate (c.f. Fig.~\ref{fig:scatter}).

\subsection{Baselines and Experimental Setup}
\label{sec:exp_setup}

For each NAS-Bench-360 task, we compare \Algo with the following methods: DenseNAS \citep{fang2020densenas} and GAEA PC-DARTS \citep{li2021gaea}, which represent general NAS; Auto-DeepLab \citep{liu2019autodl} and AMBER \citep{zhang2021amber}, which represent specialist NAS methods for dense prediction and 1D tasks, respectively; 1D temporal convolutional network (TCN) \citep{bai2018tcn}, regular WRN, and WRN with hyperparameter tuner ASHA~\citep{li2020asha}, which represent natural NAS baselines; 
and Perceiver IO~\citep{jaegle2107}, which represents non-NAS general-purpose models. While these results are available in \citet{tu2021nb360}, we additionally add a \textsc{Baby} \Algo baseline: we run \Algo in the DARTS convolution space with $K = \{3, 5\}$ and $D=\{1, 2\}$ to study whether large kernel sizes and dilations are necessary to strong performance across-the-board.
Finally, we compare our method to the expert architectures selected by NAS-Bench-360. These models are representatives of the best that hand-crafting has to offer.

Each dataset is preprocessed and split using the NAS-Bench-360 script, with the training set being used for search, hyperparameter tuning, and retraining. To construct the multi-scale search space, we set $K$ and $D$ according to the rules in Section~\ref{sec:methodslast}. 
We use the default SGD optimizer for the WRN backbone and fix the learning rate schedule as well as the gradient clipping threshold for every task. The entire \Algo pipeline can be run on a single NVIDIA V100 GPU, which is also the system that we use to report the runtime cost. Full experimental details can be found in the Appendix.

We evaluate the performance of all competing methods following the NAS-Bench-360 protocol. We first report results of the target metric for each task by running the model of the \textit{last} epoch on the test data. Then, we report aggregate results via \emph{performance profiles}~\citep{dolan2002profiles}, a technique that considers both outliers and small performance differences to compare methods across multiple tasks robustly. In such plots, each curve represents one method. The $\tau$ on the $x$-axis denotes the fraction of tasks on which a method is no worse than a $\tau$-factor from the best.

\subsection{Results and Discussion}

\renewcommand{\res}[2]{#1}
\begin{table}[t]
\large
	\centering
		\caption{\small
			Error rates (lower is better) on NAS-Bench-360 tasks across diverse application domains and problem dimensions (the last three problems are 1D and the rest are 2D). \Algo beats \textit{all the other NAS methods} on 7/10 tasks and exceeds hand-designed expert models on 7/10 tasks.
			 Scores of \Algo are averaged over three trials. Scores of the baselines are from \citet{tu2021nb360}. 
			See Table~\ref{table:accwitherror} in the Appendix~\ref{appendix:accwitherrors} for standard deviations.
		}
	    \vspace{0.2cm}
		\label{table:acc}
\resizebox{1.0\textwidth}{!}{
		\begin{tabular}{lcccccccccc}
			\toprule
			 & {\scriptsize CIFAR-100} & {\scriptsize Spherical}  & {\scriptsize Darcy Flow} & {\scriptsize PSICOV} & {\scriptsize Cosmic} & {\scriptsize NinaPro} & {\scriptsize FSD50K}  & {\scriptsize ECG} & {\scriptsize Satellite} & {\scriptsize DeepSEA} \\
			 & {\scriptsize 0-1 error (\%) } & {\scriptsize 0-1 error (\%)}& {\scriptsize relative $\ell_2$ } & {\scriptsize MAE$_8$}  & {\scriptsize 1-AUROC} & {\scriptsize 0-1 error (\%)} &{\scriptsize 1-mAP } & {\scriptsize 1-F1 } & {\scriptsize 0-1 error (\%)}  & {\scriptsize 1- AUROC  } \\
			\midrule
			Expert &\textbf{\res{19.39}{0.20}}  & \res{67.41}{0.76}  & \res{0.008}{0.001} & \res{3.35}{0.14} & \textbf{\res{0.13}{0.01}} & \res{8.73}{0.9}  & \res{0.62}{0.004}  & \textbf{\res{0.28}{0.00} } & \res{19.8}{0.00} & \res{0.30}{0.24} \\ 
			
            \midrule
		WRN  &\res{23.35}{0.05}& \res{85.77}{0.71}& \res{0.073}{0.001}& \res{3.84}{0.053}& \res{0.24}{0.015} & \res{6.78}{0.26}& \res{0.92}{0.001}&\res{0.43}{0.01}&\res{15.49}{0.03}& \res{0.40}{0.001} \\
			TCN & - & - & - & - & - & - & - &\res{0.57}{0.005}&\res{16.21}{0.05}& \res{0.44}{0.001}
			\\
            	WRN-ASHA & \res{23.39}{0.01}  & \res{75.46}{0.40} & \res{0.066}{0.00}  &\res{3.84}{0.05} & \res{0.25}{0.021} & \res{7.34}{0.76} & \res{0.91}{0.03} &\res{0.43}{0.01}  & \res{15.84}{0.52}& \res{0.41}{0.002} \\
			DARTS-GAEA &\res{24.02}{1.92}  & \textbf{\res{48.23}{2.87}}  & \res{0.026}{0.001}  & \textbf{\res{2.94}{0.13} }& \res{0.22}{0.035} & \res{17.67}{1.39} & \res{0.94}{0.02} & \res{0.34}{0.01}& \res{12.51}{0.24} &  \res{0.36}{0.02} \\ 
			DenseNAS  & \res{25.98}{0.38} &\res{72.99}{0.95} & \res{0.10}{0.01}& \res{3.84}{0.15}& \res{0.38}{0.038} & \res{10.17}{1.31}&\res{0.64}{0.002}&\res{0.40}{0.01}&\res{13.81}{0.69}&\res{0.40}{0.001} \\
			Auto-DL & - & - &\res{0.049}{0.005}&\res{6.73}{0.73}& \res{0.49}{0.004} & - & - & - & - & - \\ 
			AMBER & - & - & - & - & - & - & - & \res{0.67}{0.015}&\res{12.97}{0.07}& \res{0.68}{0.01}\\ 
			Perceiver IO  & \res{70.04}{0.44} & \res{82.57}{0.19} & \res{0.24}{0.01} &\res{8.06}{0.06}&\res{0.48}{0.01}& \res{22.22}{1.80} & \res{0.72}{0.002} & \res{0.66}{0.01} & \res{15.93}{0.08} & \res{0.38}{0.004} \\
			
			\midrule
			\textsc{Baby} \Algo &\res{25.56}{1.37}  & \res{63.45}{0.88}  & \res{0.016}{0.002}  & \res{3.94}{0.54} & \res{0.16}{0.007} &\res{8.28}{0.62}  & \res{0.62}{0.01}  & \res{0.37}{0.001}  & \res{13.29}{0.108} & \res{0.37}{0.017} \\
			
			\Algo &\res{24.37}{0.81}  & \res{71.28}{0.68}  & \textbf{\res{0.0079}{0.002}}  & \res{3.30}{0.16} & \res{0.19}{0.02} &\textbf{\res{6.60}{0.33}} & \textbf{\res{0.60}{0.008} } & \res{0.32}{0.007}  & \textbf{\res{12.28}{0.5} }& \textbf{\res{0.28}{0.013}} \\
	    \bottomrule
		\end{tabular}}
	\vspace{-1mm}
\end{table}

\begin{figure}[H]
  \centering
 \includegraphics[width=0.45\textwidth]{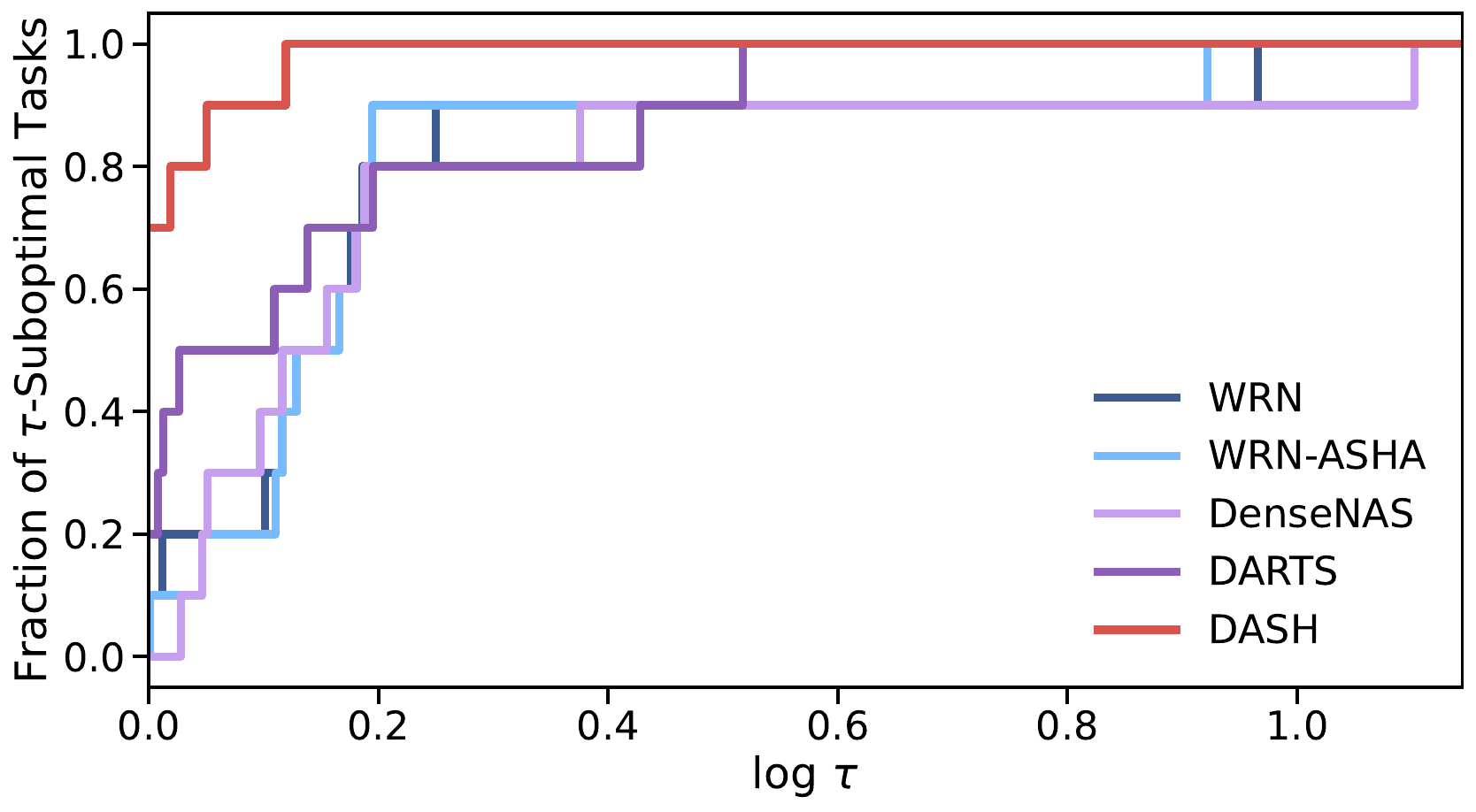}
	        \label{fig:expprofile}
	        \caption{\small
	    Performance profiles of general NAS methods and \Algo on NAS-Bench-360. 
	    \Algo{} being far in the top left corner indicates it is rarely suboptimal and is often the best.
	    }
	    \label{fig:profile}
\end{figure}

\begin{figure}[H]
  \centering
 \includegraphics[width=0.46\textwidth]{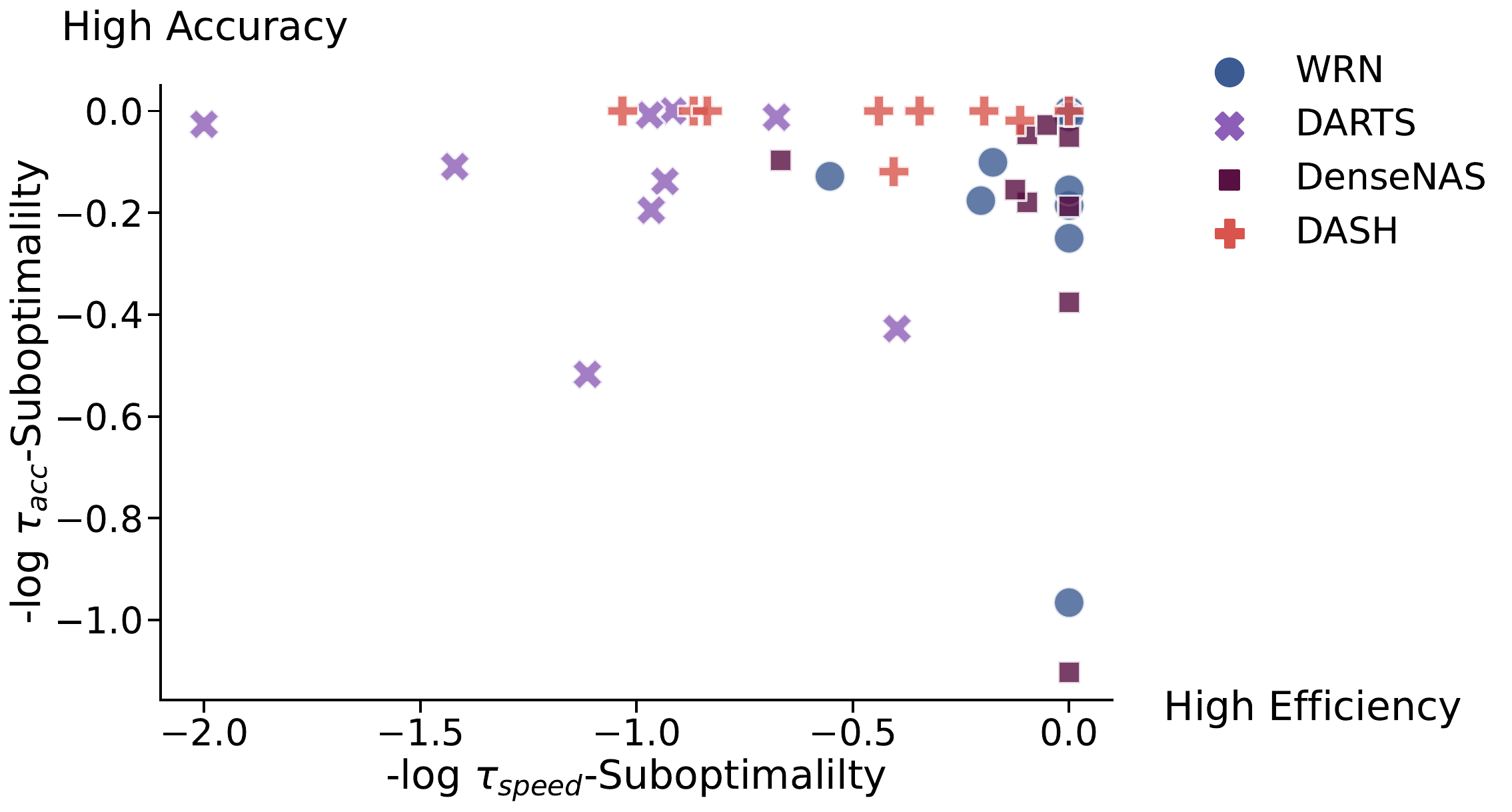}
	 \caption{\small
	        Comparing $-\log\tau$-suboptimality of speed vs.  accuracy  on all tasks.
	        \Algo{}'s concentration in the top right corner indicates its strong efficacy-efficiency trade-offs relative to the other methods.
	       \label{fig:scatter} 
	 }
\end{figure}

We present the accuracy results for each task in Table~\ref{table:acc} and the performance profiles in Fig.~\ref{fig:profile}. Fig.~\ref{fig:profile} clearly demonstrates that \Algo is superior to other competing methods in terms of aggregate performance.  In particular, it ranks first among all \textit{automated} models for 7/10 tasks, among all \textit{expert} models for 7/10 tasks, and performs favorably when considering both accuracy and efficiency as shown in Fig.~\ref{fig:scatter}. In addition, Table~\ref{table:hours} shows that \Algo outperforms DARTS in speed for all 10 tasks (in several cases by an order of magnitude), and attains comparable efficiency with training vanilla WRNs for 6/10 tasks (full-pipeline time is less than or about twice as long as the WRN training time). 
In the following, we provide a detailed analysis of the experimental results.

\paragraph{\Algo dominates automated methods.} Compared to other automated methods, \Algo has a clear advantage in accuracy. Even for tasks where it does not beat the expert, e.g. ECG, \Algo's performance is significantly better than other AutoML methods. It also outperforms specialist methods Auto-DL and AMBER on dense prediction and 1D tasks, respectively. Although DARTS does best on CIFAR-100 (the task for which it was designed), Spherical, and PSICOV, it is the worst on NinaPro and FSD50K. Note that the underperformance of \Algo on CIFAR-100 relative to WRN (and on Spherical and Cosmic relative to \textsc{Baby} \Algo) suggests suboptimality of the gradient descent optimization procedure but not of the operation space, since WRN and \textsc{Baby} \Algo are contained in our search space. This indicates a future direction to improve optimization in the \Algo search space.

\begin{multicols}{2}

\end{multicols}

\paragraph{\Algo dominates expert architectures.}
While the degree of sophistication of the expert networks varies task by task, the performance of \Algo{} on tasks such as Darcy Flow suggests that it is capable of competing with highly specialized networks, e.g., Fourier Neural Operator \citep{li2021fno} for PDE solving. These results imply that \Algo, and more generally the strategy of equipping backbone networks with  task-specific kernels, is a promising approach for tackling model development in new domains. Meanwhile, \Algo  consistently outperforms Perceiver IO which represents non-automated general-purpose models. We speculate that the poor performance of Perceiver IO is because  it is developed on more sophisticated  natural language and multi-modal reasoning tasks and training Transformers is generally difficult.

\begin{wraptable}{r}{0.49\textwidth}
\vspace{-5mm}
		\caption{\small Full-pipeline runtime in GPU hours for NAS-Bench-360 (PSICOV results are omitted due to a discrepancy in the implementation of data loading). \Algo is consistently faster than DARTS, and it is less than a factor of two slower than simply training a WRN on six of the ten tasks.
		 DenseNAS is fast but its accuracy is far less impressive. XD
		 is too expensive to be applied to tasks other than CIFAR-100 \citep{roberts2021xd}.
		 }
		\label{table:hours}
		\centering
		\small
		\resizebox{0.5\textwidth}{!}{\begin{tabular}{lcccc}
			\toprule
			Task     &  DARTS  &  DenseNAS &  WRN & \Algo\\ 
			\midrule
			CIFAR-100 & 9.5 & 2.5& 2& 2.5\\
			\midrule 
			Spherical  & 16.5 & 2.5& 2& 5\\
			\midrule
			Darcy Flow & 6.5 & 0.5& 0.5 &5.3\\ 
			\midrule 
			Cosmic & 21.5& 2.5& 4& 6.8\\ 
			\midrule
			NinaPro  & 0.5 & 0.2& 0.2& 0.3\\ 
			\midrule 
			FSD50K &37&4.5&4& 29 \\
			\midrule 
			ECG &140&6.5&5&1.3\\
			\midrule 
			Satellite &28&3&4.5&6.5 \\
			\midrule 
			DeepSEA &39.5&2&1.5&10 \\
			\bottomrule
		\end{tabular}}
	\end{wraptable}
	
\paragraph{Large kernels are needed.}
We also ablate the large-$k$-large-$d$ design of the search space by comparing \Algo with \textsc{Baby} \Algo. We hypothesize that for the same task, a small performance gap between the two methods would indicate that small kernels suffice for extracting local features, whereas a major degradation in the quality of the \textsc{Baby} \Algo model can imply that the task needs global modeling. Consequently, datasets such as Darcy Flow and ECG provide compelling evidence that kernels with large receptive fields play an important role in solving real-life problems and further back up the design of our multi-scale convolutional search space. An example of the series of WRN kernels found by \Algo on Darcy Flow is: {\small $\Conv_{5,3}\rightarrow\Conv_{3,1}\rightarrow\Conv_{3,1}\rightarrow\Conv_{3,15}\rightarrow\Conv_{7,15}\rightarrow\Conv_{9,7}\rightarrow\Conv_{9,7}\rightarrow\Conv_{3,7}\rightarrow\Conv_{5,7}\rightarrow\Conv_{5,15}\rightarrow\Conv_{9,7}\rightarrow\Conv_{3,7}\rightarrow\Conv_{7,7}$}.  We can see that large kernels are indeed selected during search. More visualizations can be found in Appendix~\ref{appendix:fig}.

\paragraph{\Algo is computationally efficient.}
In addition to prediction quality, we also care about the efficiency of model selection. Table~\ref{table:hours} provides the combined search and retraining time in GPU hours for DARTS, DenseNAS, and \Algo, as well as the training time for vanilla WRN 16-4 without hyperparameter tuning (baseline results are taken from \citet{tu2021nb360}). We also present the breakdown of \Algo's full-pipeline runtime in Appendix~\ref{sec:appendix:time}. A key observation is that the cost of \Algo is consistently below DARTS' on all tasks and is similar to training a simple CNN for more than half of them. Although DenseNAS stands out by speed, its practical performance is less impressive.

In Fig.~\ref{fig:scatter}, we visualize the trade-off between efficiency and effectiveness for each method and task combination. 
Evidently, \Algo takes an important step towards bridging the gap between the efficiency of DARTS and the expressivity of XD in NAS. The fact that \Algo can be trained at a low cost testifies that we need not sacrifice efficiency for adding more operations. In fact, we have actually shown that \Algo is \textit{both} faster \textit{and} more effective than DARTS in many tasks. 

\paragraph{\Algo works with other backbones and is backward compatible.}
Lastly, in addition to the Wide ResNet backbone and NAS-Bench-360 tasks, we have also verified the efficacy of \Algo on other backbones including TCN \citep{bai2018tcn} and ConvNeXt \citep{Liu2022ACF}, and on large-scale datasets including ImageNet in Appendix~\ref{appendix:additional}. In particular, \Algo is able to achieve a 1.4\% increase in top-1 accuracy for ImageNet on top of the ConvNeXt backbone. As the latter was itself developed in part via manual tuning of the kernel size, this means that \Algo outperforms human hand-tuning on ImageNet. These results show that \Algo is backbone-agnostic and also works well with computer vision tasks, making it backward compatible with the original use cases of CNNs.

\subsection{Limitations and Future Work}
\label{sec:impact}

There are several open problems which we leave for future work. First, it is beneficial to study why certain kernel patterns are chosen, as the selected operations can hint us at the intrinsic properties of the datasets. Second, one can improve upon \Algo, e.g., by including non-square convolutions for 2D problems, using a better optimization algorithm, or developing techniques that further reduces the memory usage of performing a forward architecture search pass.
One could also construct a more comprehensive search space with high-level operators such as  self-attention \citep{vaswani2017attention}.

Meanwhile, although this paper focuses on NAS, which is an alternative to fine-tuning pretrained models, the aggregated convolution can be a plug-and-play module for algorithms that search for large-scale models. For instance, many Transformer models still depend on convolutions for feature extraction and transformation, and their performance relies on the quality of the embedded features. Since \Algo is applicable to any architecture with a convolutional layer, it can be helpful for such models, including Vision Transformer with a convolutional patching layer \citep{dosovitskiy2021an}, Deformable Transformer with a ResNet embedder \citep{Zhu2021DeformableDD}, Swin Transformer with a convolutional decoder \citep{Liu2021SwinTH}, and many others.

\paragraph{Societal Impact}
This paper calls for ML community’s attention to less-studied application domains  and moves towards truly democratizing machine learning in real life. In terms of broader societal impact, our work can exert a positive influence as it  contributes to NAS efficiency and reduces the computational burden on AutoML end-users. However, lowering the barrier for applying ML to a wide range of tasks necessarily comes with the risk of misuse. Hence, it is imperative to develop NAS methods with privacy, safety, and fairness guarantees.

\section{Conclusion}
\label{sec:conclusion}
In this paper, we argue that a crucial goal of NAS is to discover accessible models for diverse tasks. To this end, we propose \Algo, which efficiently searches for convolution patterns and integrates them into existing backbones. \Algo overcomes the computational limitations of differentiable NAS and obtains high-quality models with accuracy comparable to or better than that of the handcrafted networks on many tasks. Our experiments show that convolution can be a universal operator for many under-explored areas. \Algo is also a promising step towards developing general-purpose models with more complicated structures.

\section*{Acknowledgments}
We thank Maria-Florina Balcan, Nicholas Roberts, and Renbo Tu for providing useful feedback. This work was supported in part by DARPA FA875017C0141, the National Science Foundation grants IIS1705121, IIS1838017, IIS2046613 and IIS-2112471, an Amazon Web Services Award, a Facebook Faculty Research Award, funding from Booz Allen Hamilton Inc., a Block Center Grant, and a Facebook Fellowship Award. Any opinions, findings and conclusions or recommendations expressed in this material are those of the author(s) and do not necessarily reflect the views of any of these funding agencies. 

%%%%%%%%%%%%%%%%%%%%%%%%%%%%%%%%%%%%%%%%%%%%%%%%%%%%%%%%%%%%
\newpage
\bibliographystyle{unsrtnat}
\bibliography{refs}

\newpage
\section*{Checklist}

\begin{enumerate}

\item For all authors...
\begin{enumerate}
  \item Do the main claims made in the abstract and introduction accurately reflect the paper's contributions and scope?
    \answerYes{}
  \item Did you describe the limitations of your work?
    \answerYes{}
  \item Did you discuss any potential negative societal impacts of your work?
    \answerYes{} Our work aims to democratize machine learning in diverse real-life applications. However, reducing the barrier for deploying accurate ML models on a diverse range of tasks may introduce unintended consequences associated with safety/fairness/privacy issues. We have addressed this point in Section~\ref{sec:impact}.
  \item Have you read the ethics review guidelines and ensured that your paper conforms to them?
    \answerYes{}
\end{enumerate}

\item If you are including theoretical results...
\begin{enumerate}
  \item Did you state the full set of assumptions of all theoretical results?
    \answerYes{}
        \item Did you include complete proofs of all theoretical results?
    \answerYes{} Complexity analysis for Table~\ref{table:complexity} is provided in Appendix~\ref{sec:appendix:analysis}.
\end{enumerate}

\item If you ran experiments...
\begin{enumerate}
  \item Did you include the code, data, and instructions needed to reproduce the main experimental results (either in the supplemental material or as a URL)?
    \answerYes{}
  \item Did you specify all the training details (e.g., data splits, hyperparameters, how they were chosen)?
    \answerYes{}
        \item Did you report error bars (e.g., with respect to the random seed after running experiments multiple times)?
    \answerYes{} Due to limited space, we present the error bars in Appendix~\ref{appendix:accwitherrors}.
        \item Did you include the total amount of compute and the type of resources used (e.g., type of GPUs, internal cluster, or cloud provider)?
    \answerYes{}
\end{enumerate}

\item If you are using existing assets (e.g., code, data, models) or curating/releasing new assets...
\begin{enumerate}
  \item If your work uses existing assets, did you cite the creators?
    \answerYes{} 
  \item Did you mention the license of the assets?
    \answerYes{} See Table~\ref{table:taskinfo}.
  \item Did you include any new assets either in the supplemental material or as a URL?
    \answerYes{}
  \item Did you discuss whether and how consent was obtained from people whose data you're using/curating?
    \answerNA{} The data we used are open-sourced and our usage comply with
their terms.
  \item Did you discuss whether the data you are using/curating contains personally identifiable information or offensive content?
    \answerNA{}  The data we used do not include any personally identifiable information.
\end{enumerate}

\item If you used crowdsourcing or conducted research with human subjects...
\begin{enumerate}
  \item Did you include the full text of instructions given to participants and screenshots, if applicable?
    \answerNA{}
  \item Did you describe any potential participant risks, with links to Institutional Review Board (IRB) approvals, if applicable?
    \answerNA{}
  \item Did you include the estimated hourly wage paid to participants and the total amount spent on participant compensation?
    \answerNA{}
\end{enumerate}

\end{enumerate}

%%%%%%%%%%%%%%%%%%%%%%%%%%%%%%%%%%%%%%%%%%%%%%%%%%%%%%%%%%%%
\newpage
\appendix

% !TEX root = main.tex

\section{Appendix}
\subsection{Term Clarification}
\label{appendix:clarification}
Since we compare with a variety of methods in the paper, here we clarify some of the terms we use. 
\begin{table}[!ht]
\centering
		\begin{tabular}{cc}
			\toprule
			 what we say & what we are referring to \\
			\midrule
			Best-Automated (Fig.~\ref{fig:introa})& WRN, WRN-ASHA, DARTS, DenseNAS, Auto-DL, AMBER \\
			Hand-Designed (Fig.~\ref{fig:introa}) & Expert architectures in Table~\ref{table:taskinfo}\\
			AutoML & WRN-ASHA, DARTS, DenseNAS, Auto-DL, AMBER \\
			NAS & DARTS, DenseNAS, Auto-DL, AMBER\\
			WRN & WRN without hyperparameter tuning\\
			\bottomrule
		\end{tabular}
\label{table:clarification}
\end{table}
\subsection{Asymptotic Analysis}\label{sec:appendix:analysis}

In this section we outline the runtime analysis used to populate the asymptotic complexities in Table~\ref{table:complexity}.
All three methods in the table---\textsl{mixed-results}, \textsl{ mixed-weights}, and \Algo---are computing the following weighted sum of convolutions:
\begin{align}
\begin{split}
\AggC_{K,D} (\*x) := \sum_{k\in K}\sum_{d\in D} \alpha_{k,d}\cdot \Conv(\*w_{k,d})(\*x).
\end{split}
\end{align}
We consider 1D inputs $\*x$ with length $n$ and $c_{in}$ input channels; the convolutions have $c_{out}$ output channels.
We view $\Conv(\*w_{k,d})(\*x)$ as having the naive complexity $c_{in}c_{out}kn$ since the deep learning frameworks use the direct (non-Fourier) algorithm.
\textsl{mixed-results} computes the sum directly, which involves (1) applying one convolution of each size $k$ and dilation to $\*x$ at a cost of $c_{in}c_{out}kn$ MULTs and ADDs each for a total cost of $c_{in}c_{out}\bar Kn$, (2) scalar-multiplying the outputs at a cost of $c_{out}|K||D|n$ MULTs, and (3) summing the results together at a cost of $c_{out}|K||D|n$ ADDs.
\textsl{mixed-weights} instead (1) multiplies all kernels by their corresponding weight at a cost of $c_{in}c_{out}\bar K$ MULTs, (2) zero-pads the results to the largest effective kernel size $\bar D$ and adds them together at a cost of $c_{in}c_{out}|K||D|\bar D$ ADDs, and (3) applies the resulting $\bar D$-size convolution to the input at a cost of $c_{in}c_{out}\bar Dn$ MULTs and ADDs.
Finally, \Algo also (1) does the first two steps of \textsl{ mixed-weights} at a cost of $c_{in}c_{out}\bar K$ MULTs and $c_{in}c_{out}|K||D|\bar D$ ADDs but then (2) pads the resulting $\bar D$-size convolution to size $n$ and applies an FFT at a cost of $\BigO(c_{in}c_{out}n\log n)$ MULTs and ADDs, (3) applies an FFT to $\*x$ at a cost of $\BigO(c_{in}n\log n)$, (4) element-wise multiplies the transformed filters by the inputs at a cost of $c_{in}c_{out}n$ MULTs, (5) adds up $c_{in}$ results for each of $c_{out}$ output channels at a cost of $c_{in}c_{out}$ MULTs, and (6) applies an iFFT to the result at a cost of $\BigO(c_{out}n\log n)$.

\subsection{Experiment Details for Fig.~\ref{fig:time1} and 
\label{sec:appendix:mnist}
Fig.~\ref{fig:time2}}
For the speed tests, we work with the Sequential MNIST dataset, i.e., the 2D $28 \times 28$ images are stretched into 1D with length $784$. We zero pad or truncate the input to generate data with different input size $n$. The backbone is 1D WRN with the same structure as introduced in Section~\ref{sec:methods}. The batch size is $128$. We run the workflow on a single NVIDIA V100 GPU. The timing results reported are the $\log_{10}($combined forward and backward pass time for one search epoch$)$.

In Fig.~\ref{fig:time1}, we study how the size of our multi-scale convolution search space affects the runtimes of \textsl{mixed-results}, \textsl{mixed-weights}, and \Algo for $n=1000$ (zero-padded MNIST). We define $K = \{3+2(p-1) | 1\leq p \leq c\}$, $D = \{2^q-1 | 1\leq q\leq c\}$ and varies $c$ from $1$ to $7$. Consequently, the number of operations included in the search space grows from $1$ to $49$. 

In Fig.~\ref{fig:time2}, we study how the input size affects the runtimes of the three methods. We fix $K = \{3, 5, 7, 9, 11\}$, $D = \{1, 3, 7, 15, 31\}$ and vary $n$ from $2^5$ to $2^{12}$.

\newpage
\subsection{Information About Tasks in NAS-Bench-360}

\begin{table}[h!]
	\caption{Information about evaluation tasks in NAS-Bench-360 \citep{tu2021nb360}.}
	\vspace{2mm}
	\centering
	\resizebox{1\textwidth}{!}{\begin{tabular}{lllllll}
		\toprule
		Task name   & \# Data  & Data dim. & Type & License &  Learning objective &Expert arch.\\
		\midrule
		\multirow{2}{*}{CIFAR-100} & \multirow{2}{*}{60K}  & \multirow{2}{*}{2D} & \multirow{2}{*}{Point} & \multirow{2}{*}{CC BY 4.0}& \multirow{2}{*}{Classify natural images into 100 classes} & DenseNet-BC \\
		&&& && & \citep{Huang2017DenselyCC}\\
		\midrule
		\multirow{2}{*}{Spherical} & \multirow{2}{*}{60K}  &  \multirow{2}{*}{2D} & \multirow{2}{*}{Point} &\multirow{2}{*}{CC BY-SA}&  Classify spherically projected images   &S2CN \\
&& &&& into 100 classes &\citep{Cohen2018SphericalC}\\
		\midrule
		\multirow{2}{*}{NinaPro} & \multirow{2}{*}{3956}  &  \multirow{2}{*}{2D}& \multirow{2}{*}{Point} &\multirow{2}{*}{CC BY-ND}&  Classify sEMG signals into 18 classes   & Attention Model\\
& &&&&corresponding to hand gestures &\citep{Josephs2020sEMGGR} \\
		\midrule
		\multirow{2}{*}{FSD50K} & \multirow{2}{*}{51K} &  \multirow{2}{*}{2D} & 
		Point & \multirow{2}{*}{CC BY 4.0}&Classify sound events in log-mel & VGG\\ 
& &&(multi-label)&  &spectrograms with 200 labels &\citep{Fonseca2021FSD50KAO} \\
		\midrule
		\multirow{2}{*}{Darcy Flow} & \multirow{2}{*}{1100}  & \multirow{2}{*} {2D} & \multirow{2}{*}{Dense} & \multirow{2}{*}{MIT}& Predict the final state of a fluid from its  & FNO\\
&& &&& initial conditions &\citep{li2021fno}\\	
		\midrule
		\multirow{2}{*}{PSICOV} & \multirow{2}{*}{3606} &  \multirow{2}{*}{2D}  & \multirow{2}{*}{Dense} &\multirow{2}{*}{GPL}&  Predict pairwise distances between resi- &DEEPCON \\
&&& && duals from 2D protein sequence features & \citep{10.1093/bioinformatics/btz593}\\

		\midrule
\multirow{2}{*}{Cosmic} & \multirow{2}{*}{5250} &  \multirow{2}{*}{2D}  & \multirow{2}{*}{Dense} & \multirow{2}{*}{Open License}& Predict propablistic maps to identify cos- &deepCR-mask \\
&& &&& mic rays in telescope images  &\citep{Zhang2019deepCRCR} \\	

		\midrule 
\multirow{2}{*}{ECG} & \multirow{2}{*}{330K} &  \multirow{2}{*}{1D}  & \multirow{2}{*}{Point} & \multirow{2}{*}{ODC-BY 1.0} &  Detect atrial cardiac disease from & ResNet-1D\\
&& &&& a ECG recording (4 classes) &\citep{Hong2020HOLMESHO} \\

		\midrule 
\multirow{2}{*}{Satellite} & \multirow{2}{*}{1M} &  \multirow{2}{*}{1D}  & \multirow{2}{*}{Point}&
\multirow{2}{*}{GPL 3.0}  &  Classify satellite image pixels' time   & ROCKET \\
&& &&&  series into 24 land cover types  &\citep{Dempster2020ROCKETEF} \\

		\midrule
\multirow{2}{*}{DeepSEA} & \multirow{2}{*}{250K} &  \multirow{2}{*}{1D} & 
Point & \multirow{2}{*}{CC BY 4.0} &Predict chromatin states and binding  &
\multirow{2}{*}{} DeepSEA\\ 
&&&(multi-label)&  &states of RNA sequences (36 classes)  &\citep{Zhou2015PredictingEO} \\
		\bottomrule
	\end{tabular}}
\label{table:taskinfo}
\end{table}

\subsection{Evaluation of \Algo on NAS-Bench-360}
\subsubsection{Backbone Network Structure}
\paragraph{2D Tasks}
We use the Wide ResNet 16-4 \citep{zagoruyko2016wideresnet} as the backbone for all 2D tasks. The original model is made up of 16 $3\times 3$ conv followed by $6$ WRN blocks with the following structure ($i\in\{1, 2, 3, 4, 5, 6\}$ indicates the block index):
\begin{table}[ht]
\centering
\small
\begin{tabular}{|c|c|}
\hline
\multicolumn{2}{|c|}{BatchNorm, ReLU}  \\ \hline
Conv 1 & $16 \times 4 \times ((i+1)//2) $ $(k=3, d=1)$ filters, ReLU\\ \hline
Dropout & dropout rate $p$  \\ \hline
\multicolumn{2}{|c|}{BatchNorm, ReLU}  \\ \hline
Conv 2 &   $16 \times 4 \times ((i+1)//2) $ $(k=3, d=1)$  filters , stride = $(i+1)//2$, ReLU \\ \hline
\multicolumn{2}{|c|}{Add residual (apply point-wise conv first if $c_{in}\neq c_{out}$)}  \\ \hline
\end{tabular}
\end{table}

The output block consists of a BatchNorm layer, a ReLU activation, a linear layer, and a final activation layer which we modify according to the task learning objective, e.g., log softmax for classification and sigmoid for dense prediction. We set $p=0$ in search and tune $p$ as a hyperparameter for retraining. We use the WRN code provided here: \url{https://github.com/meliketoy/wide-resnet.pytorch}.

\newpage
\paragraph{1D Tasks}
We use the 1D WRN \citep{IsmailFawaz2020InceptionTimeFA} as the backbone for all 1D tasks. The model is made up of $3$ residual blocks with the following structure:
\begin{table}[h!]
\centering
\small
\begin{tabular}{|c|c|}
\hline
Conv 1 & $c_{out}$ $(k=8, d=1)$ filters\\ \hline
Dropout & dropout rate $p$  \\ \hline
\multicolumn{2}{|c|}{BatchNorm, ReLU}  \\ \hline
Conv 2 & $c_{out}$ $(k=5, d=1)$ filters\\ \hline
Dropout & dropout rate $p$  \\ \hline
\multicolumn{2}{|c|}{BatchNorm, ReLU}  \\ \hline
Conv 3 & $c_{out}$ $(k=3, d=1)$ filters\\ \hline
Dropout & dropout rate $p$  \\ \hline
\multicolumn{2}{|c|}{BatchNorm, ReLU}  \\ \hline
\end{tabular}
\end{table}

In the original architecture, $c_{out} = 64$. We set $c_{out}$ to $\min(4^{\text{num\_classes}//10 + 1}, 64)$ to account for simpler tasks with fewer class labels. The output block consists of a linear layer and a activation layer which we modify according to the task learning objective, e.g., log softmax for classification and sigmoid for dense prediction. We set $p=0$ in search and tune $p$ as a hyperparameter for retraining. We use the 1D WRN code provided here: \url{https://github.com/okrasolar/pytorch-timeseries}.
 
\subsubsection{\Algo Pipeline Hyperparameters}
\label{appendix:2dhyperparam}
\paragraph{Search}
\begin{itemize}[leftmargin=0.6cm]
\item Epoch: $100$
    \item Optimizer: \texttt{SGD(momentum=0.9, nesterov=True, weight\_decay=5\text{e-}4)} for both model weights and architecture parameters
    \item Model weight learning rate: $0.1$ for point prediction tasks, $0.01$ for dense tasks
    \item Architecture parameter learning rate: $0.05$ for point prediction tasks, $0.005$ for dense tasks
    \item Learning rate scheduling: decay by $0.2$ at epoch $60$
    \item Gradient clipping threshold: $1$
    \item Softmax temperature: $1$
    \item Subsampling ratio: $0.2$
\end{itemize}

\paragraph{Hyperparameter tuning}
\label{appendix:hyperparamconfig}
\begin{itemize}[leftmargin=0.6cm]
\item Epoch: $80$
\item Configuration space:
\begin{itemize}
    \item Learning rate: $\{1\text{e-}1, 1\text{e-}2, 1\text{e-}3\}$ 
    \item Weight decay: $\{5\text{e-}4, 5\text{e-}6\}$
    \item Momentum: $\{0.9, 0.99\}$ 
    \item Dropout rate: $\{0, 0.05\}$
\end{itemize}
\end{itemize}

\paragraph{Retraining}
\begin{itemize}[leftmargin=0.6cm]
    \item Epoch: $200$
    \item Learning rate scheduling: for 2D tasks, decay by $0.2$ at epoch $60$, $120$, $160$; for 1D tasks, decay by $0.2$ at epoch $30$, $60$, $90$, $120$, $160$
\end{itemize}
\paragraph{Task-Specific Hyperparameters}

\newpage

\begin{table}[h!]
\centering
		\resizebox{1\textwidth}{!}{\begin{tabular}{cccccccc}
			\toprule
			 2D tasks & CIFAR-100 &	Spherical &	Darcy Flow&	PSICOV &	Cosmic &	NinaPro &	FSD50K  \\
			\midrule
			Batch size&64&64&10&8&4&128&128\\
			Input size & (32, 32)&(60, 60)&(85, 85)&(128, 128)&(128, 128)&(16, 52)& (96, 101)\\
			Kernel sizes $(K)$ &$\{3, 5, 7, 9\}$&$\{3, 5, 7, 9\}$&$\{3, 5, 7, 9\}$&$\{3, 5, 7, 9\}$&$\{3, 5, 7, 9\}$&$\{3, 5, 7, 9\}$&$\{3, 5, 7, 9\}$\\
			Dilations $(D)$&$\{1, 3, 7, 15\}$&$\{1, 3, 7, 15\}$&$\{1, 3, 7, 15\}$&$\{1, 3, 7, 15\}$&$\{1, 3, 7, 15\}$&$\{1, 3, 7, 15\}$&$\{1, 3, 7, 15\}$\\
			Loss $(l)$&Cross Entropy&Cross Entropy&L2& MSE& BCE w. Logits &Focal & BCE w. Logits\\
			\bottomrule
		\end{tabular}}
\label{table:2dconfig}
\end{table}

\begin{table}[h!]
\centering
		\resizebox{0.55\textwidth}{!}{\begin{tabular}{cccc}
			\toprule
			 1D tasks & Satellite & ECG &DeepSEA \\
			\midrule
			Batch size&256&1024&256\\
			Input size &46&1000&1000\\
			Kernel sizes $(K)$ &$\{3, 7, 11, 15, 19\}$&$\{3, 7, 11, 15, 19\}$&$\{3, 7, 11, 15, 19\}$\\
			Dilations $(D)$&$\{1, 3, 7,15\}$&$\{1, 3, 7, 15\}$&$\{1, 3, 7, 15\}$\\
			Loss $(l)$&Cross Entropy&Cross Entropy&BCE w. Logits\\
			\bottomrule
		\end{tabular}}
\label{table:1dconfig}
\end{table}

\subsection{Accuracy Results on NAS-Bench-360 with Error Bars}
\label{appendix:accwitherrors}
\renewcommand{\res}[2]{#1$\pm$#2}
\begin{table}[h!]
	\centering
	\begin{threeparttable}
		\caption{
			Error rates (lower is better) of \Algo and the baselines on tasks in NAS-Bench-360. Methods are grouped into three classes: non-automated, automated, and the \Algo family. Results of \Algo are averaged over three trials using the models obtained after the last retraining epoch.
			\vspace{0.1cm}
		}
		
		\label{table:accwitherror}
		\begin{tabular}{lccccc}
			\toprule
			 & CIFAR-100 & Spherical  & Darcy Flow & PSICOV & Cosmic \\
			  & 0-1 error(\%) & 0-1 error(\%)& relative $\ell_2$ & MAE$_8$  & 1-AUROC\\
			
			\midrule
			WRN  &\res{23.35}{0.05}& \res{85.77}{0.71}& \res{0.073}{0.001}& \res{3.84}{0.053}& \res{0.24}{0.015} \\
			Expert &\textbf{\res{19.39}{0.20}}  & \res{67.41}{0.76}  & \res{0.008}{0.001} & \res{3.35}{0.14} & \textbf{\res{0.13}{0.01}}   \\ 
			Perceiver IO  & \res{70.04}{0.44} & \res{82.57}{0.19} & \res{0.24}{0.01} &\res{8.06}{0.06}&\res{0.48}{0.01} \\
            \midrule
            	WRN-ASHA &  \res{23.39}{0.01}  & \res{75.46}{0.40} & \res{0.066}{0.00}  &\res{3.84}{0.05} & \res{0.25}{0.021}  \\
			DARTS-GAEA &\res{24.02}{1.92}  & \textbf{\res{48.23}{2.87}}  & \res{0.026}{0.001}  & \textbf{\res{2.94}{0.13} }& \res{0.22}{0.035}  \\ 
			DenseNAS  & \res{25.98}{0.38} &\res{72.99}{0.95} & \res{0.10}{0.01}& \res{3.84}{0.15}& \res{0.38}{0.038}\\
			Auto-DL & - & - &\res{0.049}{0.005}&\res{6.73}{0.73}& \res{0.49}{0.004}\\ 
			\midrule
			\textsc{Baby} \Algo & \res{25.56}{1.37}  & \res{63.45}{0.88}  & \res{0.016}{0.002}  & \res{3.94}{0.54} & \res{0.16}{0.007}\\
			\Algo &\res{24.37}{0.81}  & \res{71.28}{0.68}  & \textbf{\res{0.0079}{0.002}}  & \res{3.30}{0.16} & \res{0.19}{0.02} \\
			
			\toprule
			\toprule
			
			 & NinaPro & FSD50K  & ECG & Satellite & DeepSEA \\	
			  & 0-1 error (\%) & 1- mAP & 1-F1 & 0-1 error (\%) & 1-AUROC\\
			
			\midrule
			WRN  &  \res{6.78}{0.26}& \res{0.92}{0.001}&\res{0.43}{0.01}&\res{15.49}{0.03}& \res{0.40}{0.001} \\
			TCN  & -& -&\res{0.57}{0.005}&\res{16.21}{0.05}& \res{0.44}{0.001} \\
			Expert  & \res{8.73}{0.9}  & \res{0.62}{0.004}  & \textbf{\res{0.28}{0.00} } & \res{19.8}{0.00} & \res{0.30}{0.24} \\ 
Perceiver IO  &  \res{22.22}{1.80} & \res{0.72}{0.002} & \res{0.66}{0.01} & \res{15.93}{0.08} & \res{0.38}{0.004} \\
        	\midrule
			WRN-ASHA &  \res{7.34}{0.76} & \res{0.91}{0.03} &\res{0.43}{0.01}  & \res{15.84}{0.52}& \res{0.41}{0.002} \\
			DARTS-GAEA &  \res{17.67}{1.39} & \res{0.94}{0.02} & \res{0.34}{0.01}& \res{12.51}{0.24} &  \res{0.36}{0.02} \\ 
			DenseNAS  &  \res{10.17}{1.31}&\res{0.64}{0.002}&\res{0.40}{0.01}&\res{13.81}{0.69}&\res{0.40}{0.001}\\
		
			AMBER &-& -& \res{0.67}{0.015}&\res{12.97}{0.07}& \res{0.68}{0.01}\\ 	
			\midrule

			\textsc{Baby} \Algo &\res{8.28}{0.62}  & \res{0.62}{0.01}  & \res{0.37}{0.001}  & \res{13.29}{0.108} & \res{0.37}{0.017} \\
			
			\Algo &\textbf{\res{6.60}{0.33}} & \textbf{\res{0.60}{0.008} } & \res{0.32}{0.007}  & \textbf{\res{12.28}{0.5} }& \textbf{\res{0.28}{0.013}}\\
			
			\bottomrule
		\end{tabular}
	\end{threeparttable}
\end{table}

\newpage
\subsection{Runtime of \Algo on NAS-Bench-360}
\label{sec:appendix:time}
\begin{table}[h!]
\caption{Runtime breakdown for \Algo on NAS-Bench-360 tasks evaluated on a NVIDIA V100 GPU.}
\label{table:timebreakdown}
\vspace{0.1cm}
	\centering
	\begin{tabular}{lllll}
		\toprule
		Task     &  Search  &  Hyperparameter Tuning &  Retraining & Total\\ 
		\midrule
		CIFAR-100 & 1.6 & 0.15& 0.77& 2.5\\
		\midrule 
		Spherical  & 1.6 & 0.25& 3.16& 5.0\\
		\midrule
		Darcy Flow & 0.16 & 1.6& 3.5 &5.3\\ 
		\midrule
		PSICOV & 0.88 &0.64& 14& 15\\ 
		\midrule 
		Cosmic & 1.6& 0.055& 5.1& 6.8\\ 
		\midrule
		NinaPro  & 0.028 & 0.16& 0.11& 0.30\\ 
		\midrule 
		FSD50K &0.88&0.88&27& 29 \\
		\midrule 
		ECG &0.18&0.28&0.83&1.3\\
		\midrule 
		Satellite &1.8&0.4&4.3&6.5 \\
		\midrule 
		DeepSEA &0.36&1.6&8.3&10 \\
		\bottomrule
	\end{tabular}
\end{table}

\subsection{Searched Architecture Visualization}
\label{appendix:fig}
In this section, we give two example networks searched by \Algo to show that large kernel matters for diverse tasks.

\subsubsection{2D Example: Darcy Flow}
For this problem, \Algo generates a WRN 16-4 \citep{zagoruyko2016wideresnet} for retraining. The network architecture consists of several residual blocks. For instance, we can use Block$_{64, (7,1),(9,3)}$ to denote the residual block with the following structure:
\vspace{-2mm}
\begin{figure}[H]
  \centering
    \includegraphics[width=0.5\textwidth]{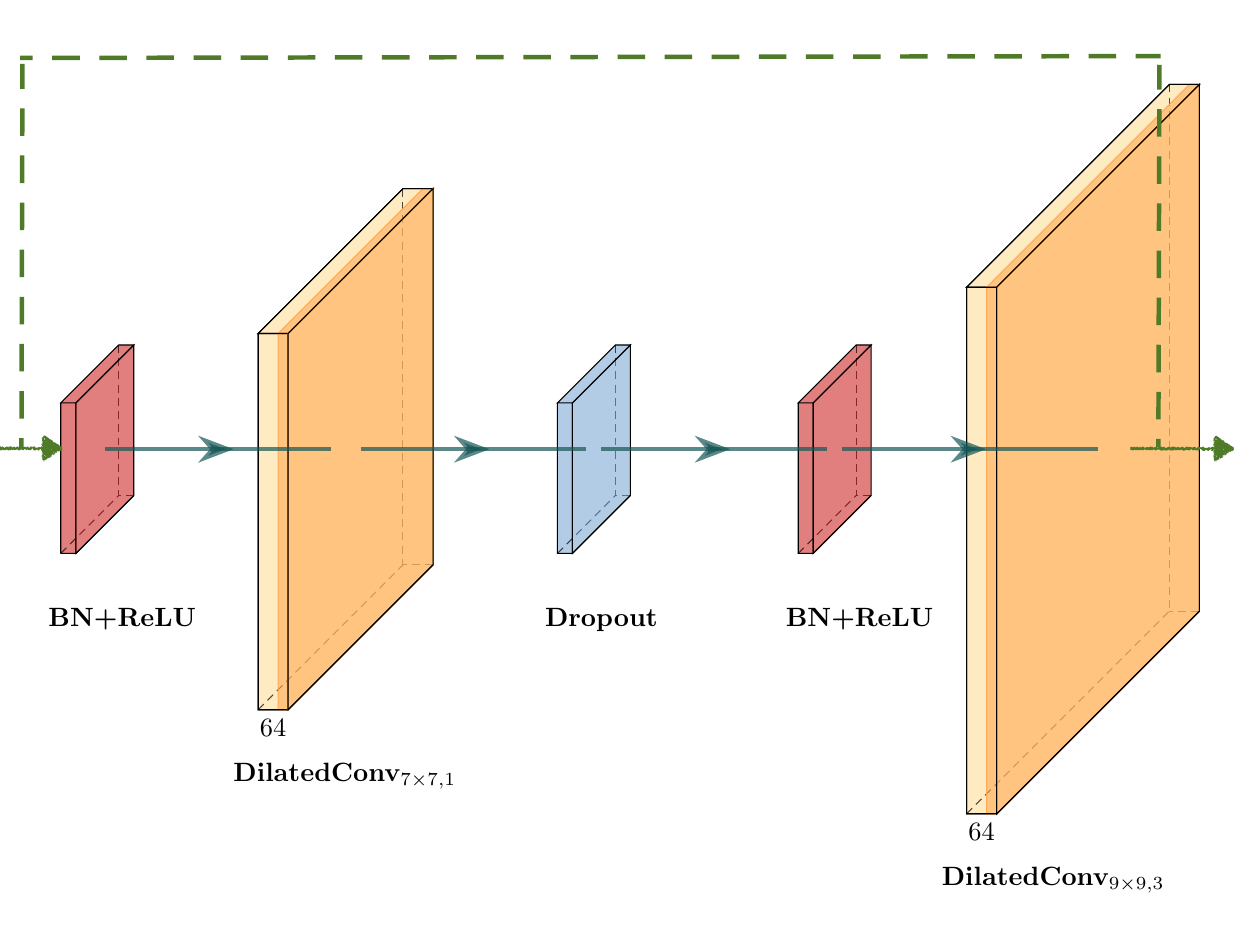}
	    \label{appendix:fig:2d1}
\end{figure}
\vspace{-2mm}
where $64$ is the output channel and BN denotes the BatchNorm layer. Note that  size of a convolutional layer in the figure is proportional to the kernel size but not the number of channels. Then, an example network produced by \Algo for Darcy Flow looks like the following:
\begin{figure}[H]
  \centering
    \includegraphics[width=1.05\textwidth]{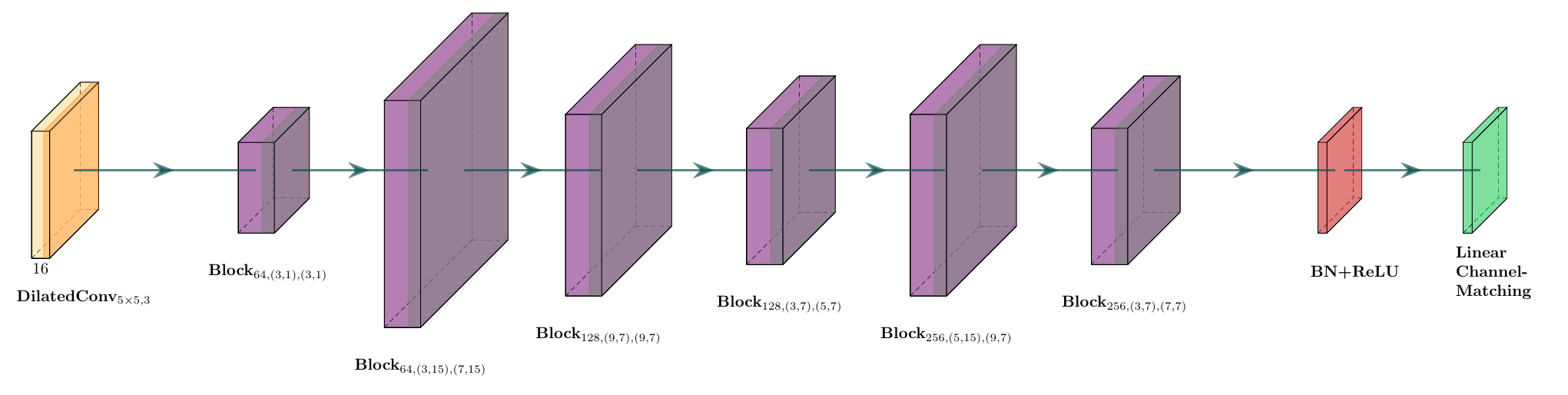}
	    \label{appendix:fig:2d2}
\end{figure}
Since Darcy Flow is a dense prediction task, the last layer is a channel-matching  (permutation+linear+permutation) layer instead of a pooling+linear layer for classification.

\subsubsection{1D Example: DeepSEA}
For this problem, \Algo generates a 1D WRN \citep{IsmailFawaz2020InceptionTimeFA} for retraining. The network architecture consists of several residual blocks. For instance, we can use Block$_{64, (3,1),(5,3),(7,5)}$ to denote the residual block with the following structure:
\vspace{-2mm}
\begin{figure}[H]
  \centering
    \includegraphics[width=\textwidth]{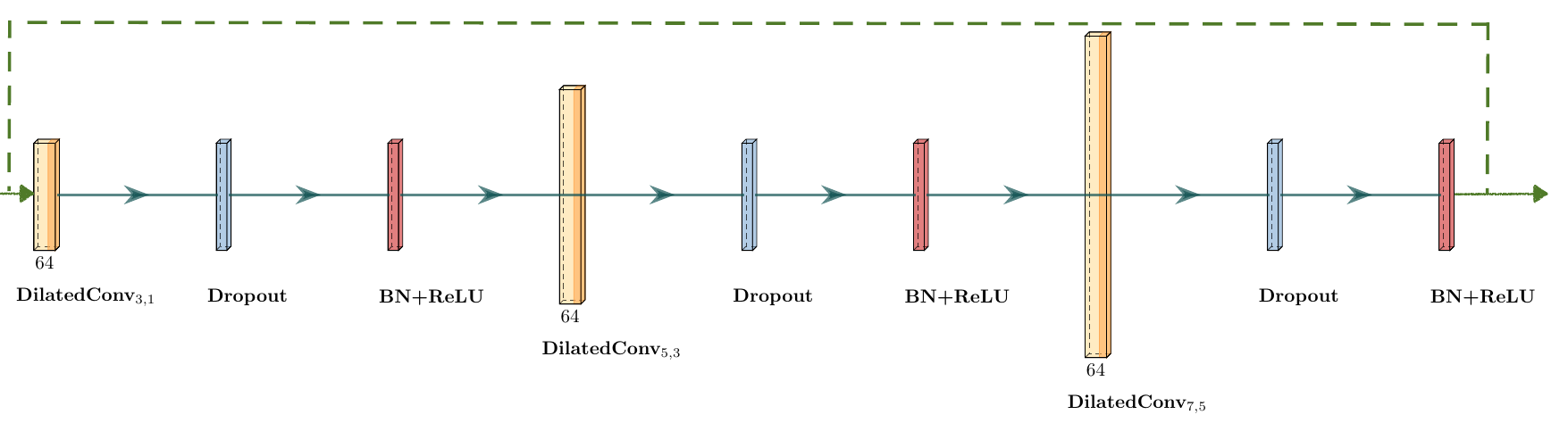}
	    \label{appendix:fig:1d1}
\end{figure}
\vspace{-2mm}
where $64$ is the output channel and BN denotes the BatchNorm layer. Then, an example network produced by \Algo for DeepSEA looks like the following:
\begin{figure}[H]
  \centering
    \includegraphics[width=0.8\textwidth]{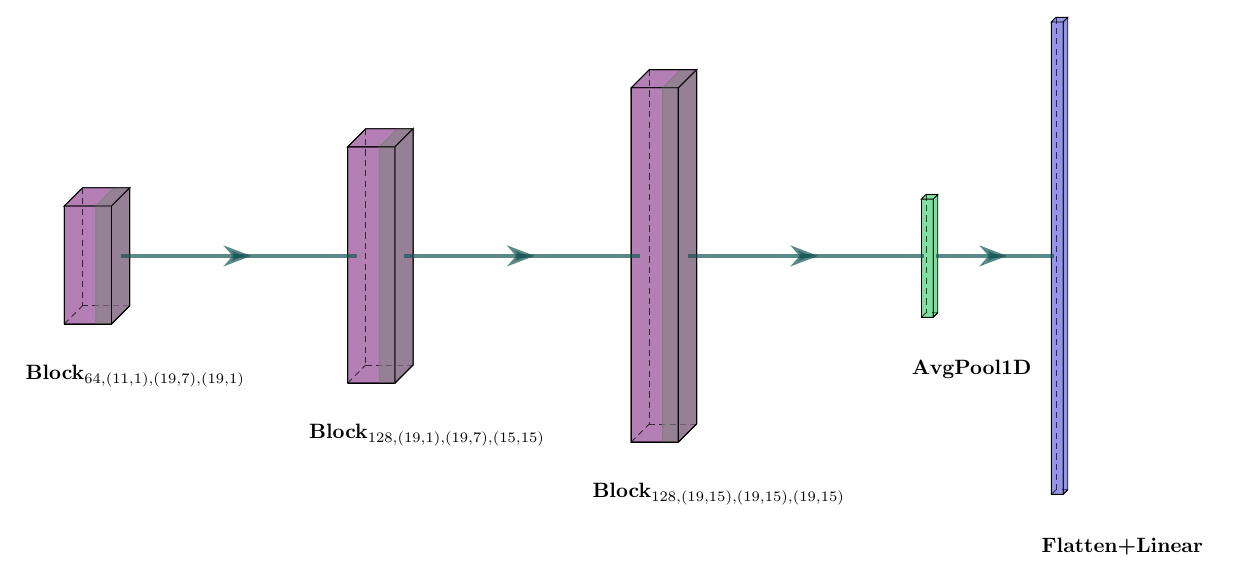}
	    \label{appendix:fig:1d2}
\end{figure}
We can see that large kernels are indeed selected during search.

\newpage
\subsection{Additional Results}
\label{appendix:additional}
\subsubsection{\Algo-TCN for NAS-Bench-360}
\Algo works for all networks with a convolutional layer, so WRNs are not the only applicable backbone. Below, we provide  the  test errors of  \Algo with the 1D Temporal Convolutional Network backbone on some 1D tasks:
\begin{table}[h!]
\vspace{-0.1cm}
\caption{Test errors for 1D NAS-Bench-360 tasks using the TCN backbone.}
\label{appendix:table:tcn}
	\centering
	\vspace{0.1cm}
	\begin{tabular}{cccc}
		\toprule
		     &  ECG & Satellite & DeepSEA \\ 
		\midrule
		Vanilla TCN & \res{0.57}{0.005}&\res{16.21}{0.05}& \res{0.44}{0.001}\\
		\Algo-TCN  & \textbf{\res{0.29}{0.004}}&\textbf{\res{12.39}{0.043}}& \textbf{\res{0.24}{0.012}}\\
		\bottomrule
	\end{tabular}
\end{table}

We did not include the results in the paper to simplify presentation. Also, using WRNs in our workflow allows us to provide a fully automated pipeline that generates decent-performing models as quickly (due to its small size) and easily (due to the code for training WRNs being easily found online) as possible for previously unexplored tasks.

\subsubsection{\Algo-ConvNeXt for ImageNet}
Though our motivation is not to compete in the crowded vision domain but to provide a general solution to less-studied domains, we show that \Algo is backward compatible with vision tasks by testing it on ImageNet-1K with two backbones of distinct scales. Our results show that \Algo generalizes to tasks with large input shape ($3\times 224\times 224$), dataset size (1.2M), and number of classes (1000). It improves the accuracy of the original models and searches efficiently regardless of the backbone used.

We used Wide ResNet 16-4 (to be consistent with our workflow) and ConvNeXt-T \citep{Liu2022ACF} (a large-scale CNN that has onpar performance with SoTA Transformers) as the backbones and performed experiments on 4 NVIDIA V100 GPUs. To demonstrate \Algo's efficiency, we first present the per-epoch search time (forward and backward time in secs) for three baselines over the search space $K=\{3,5,7,9,11\}$, $D=\{1,3,7\}$. A subset of 4096 images is used.

\begin{table}[h!]
\vspace{-0.1cm}
\caption{Time for one search epoch (forward \& backward) in seconds using different backbones.}
\vspace{0.1cm}
\label{appendix:table:imnettime}
	\centering
	\begin{tabular}{ccc}
		\toprule
		   &WRN&ConvNeXt\\
\midrule
\# param&3M&28M\\
\midrule
\Algo&151.3&80.5\\
\textsl{Mixed-weights}&705.4&300.1\\
\textsl{Mixed-results}&330.6&149.6\\
		\bottomrule
	\end{tabular}
\end{table}
We can see that \Algo's efficiency holds for both backbones. Though ConvNeXt has more parameters, it is searched faster than WRN as it has fewer conv layers and applies downsampling to the input.

Then, we report \Algo's runtime vs. the train-time of the vanilla backbone (in hours). We let \Algo search for 10 epochs with subsampling ratio 0.2. (Re)training takes 50 and 100 epochs for WRN and ConvNeXt, respectively.

\begin{table}[h!]
\vspace{-0.1cm}
\caption{Runtime breakdown for \Algo and the backbones on ImageNet-1K.}
\vspace{0.1cm}
\label{appendix:table:totaltime}
	\centering
	\begin{tabular}{ccc}
		\toprule
		   &WRN&ConvNeXt\\
\midrule
\Algo search&24&13\\
\Algo retrain&52&48\\
Backbone train &16&41\\
		\bottomrule
	\end{tabular}
\end{table}

Lastly, we report the top-1 accuracy of the searched vs. original models to show \Algo generalizes to large vision input. We trained ConvNeXt for 300 epochs.

\begin{table}[h!]
\vspace{-0.1cm}
\caption{Prediction errors (\%) for \Algo and backbones on ImageNet-1K. Backbone results are taken from \cite{Liu2022ACF}.}
\vspace{0.1cm}
\label{appendix:table:acc}
	\centering
	\begin{tabular}{ccc}
		\toprule
		   &WRN&ConvNeXt\\
\midrule
Vanilla Backbone&\res{37.56}{0.14}&\res{17.9}{0.0}\\
\Algo Searched Model&\textbf{\res{34.12}{0.21}}&\textbf{\res{16.42}{0.15}}\\
		\bottomrule
	\end{tabular}
\end{table}

In general, \Algo improves backbone performance by adopting task-specific kernels.

\end{document}